\documentclass[10pt,twocolumn,letterpaper]{article}

\usepackage[pagenumbers]{iccv} 

\usepackage[accsupp]{axessibility}
\usepackage{algorithm}  
\usepackage{algorithmicx}  
\usepackage[noend]{algpseudocode}
\usepackage{lipsum}
\usepackage{multirow}
\usepackage{multicol}
\usepackage{tabulary}
\usepackage{caption} 
\usepackage{xspace}
\usepackage{tikz}
\usepackage{lipsum}
\usepackage{subcaption} 
\usepackage{amssymb}
\usepackage{pifont}
\usepackage{wrapfig}
\usepackage{placeins}

\newcommand{\tinycodesize}{\fontsize{7.0pt}{7.0pt}\selectfont}
\newcommand*\circled[1]{\tikz[baseline=(char.base)]{
            \node[shape=circle,fill,inner sep=1pt,font=\tinycodesize] (char) {\textcolor{white}{#1}};}}
\newcommand{\oursampling}{MDPS}
\newcommand{\ours}{FastPoint}
\newcommand{\geomeanspeedup}{2.55$\times$\xspace}

\definecolor{iccvblue}{rgb}{0.21,0.49,0.74}
\usepackage[pagebackref,breaklinks,colorlinks,allcolors=iccvblue]{hyperref}

\def\paperID{6743} 
\def\confName{ICCV}
\def\confYear{2025}

\title{FastPoint: Accelerating 3D Point Cloud Model Inference via\\ Sample Point Distance Prediction}

\author{
Donghyun Lee$^{1}$ \quad Dawoon Jeong$^{1}$ \quad Jae W. Lee$^{1}$ \quad Hongil Yoon$^{1,2}$\\
$^{1}$Seoul National University \quad
$^{2}$Google\\
{\tt\small \{eudh1206, daun20211, jaewlee\}@snu.ac.kr, hongilyoon@google.com}
}

\begin{document}

\maketitle
\begin{abstract}
Deep neural networks have revolutionized 3D point cloud processing, yet efficiently handling large and irregular point clouds remains challenging. 
To tackle this problem, we introduce \ours, a novel software-based acceleration technique that leverages the predictable distance trend between sampled points during farthest point sampling. 
By predicting the distance curve, we can efficiently identify subsequent sample points without exhaustively computing all pairwise distances.
Our proposal substantially accelerates farthest point sampling and neighbor search operations while preserving sampling quality and model performance.
By integrating \ours\space into state-of-the-art 3D point cloud models, we achieve \geomeanspeedup\space end-to-end speedup on NVIDIA RTX 3090 GPU without sacrificing accuracy.

\end{abstract}
    
\section{Introduction}
\label{sec:intro}
3D point clouds have become increasingly important for representing and understanding 3D scenes, driving advancements in fields like robotics and autonomous driving. Deep neural networks~\cite{pointnet, pointnet++, assanet, fpointnet, dgcnn, randlanet, pointvector, pointmetabase, minkowski, cylinder3d, sst, flatformer, swin3d, ptr, ptr2, ptr3} have emerged as powerful tools for processing this type of data. While PointNet++~\cite{pointnet++} and its successors have improved efficiency and performance of these point cloud models, the irregular nature and growing size of point cloud data continue to impose significant computational challenges, especially for real time use cases. Operations such as Farthest Point Sampling (FPS) and neighbor search remain major bottlenecks.

While various hardware accelerators~\cite{mesorasi, crescent, quickfps, mars, dynamicfps, single_coord, fusionarch} have been explored to accelerate the models, software-only solutions are less common. Given the high cost and complexity of developing specialized accelerators, software-based acceleration is a more practical approach. Recently, several software-based techniques~\cite{edgepc, adjustable_fps} have been proposed to accelerate these models through approximation. However, such approaches often come at the cost of significant accuracy loss, limiting their practical use. 

To address this challenge, we propose \emph{\ours, a novel software-based acceleration technique with sample point distance prediction}. 
We observe that the inherent nature of FPS, which iteratively selects the point farthest from the current set of sampled points, leads to two predictable trends in the minimum distances between these points:
\begin{enumerate}
    \item \textit{Decreasing minimum distance}: As FPS makes progress, the number of remaining points decreases, forcing the algorithm to select points closer to those already sampled. This results in a smoothly decreasing curve of minimum distances between sampled points.
    \item \textit{Early structure capture}: The initial sample points tend to be the extremities of the point cloud, effectively capturing its overall shape and boundaries.
\end{enumerate}

Our proposal leverages these observations to accurately estimate the distance curve using only a few initial FPS iterations. By predicting the distance curve, we can efficiently identify subsequent sample points without computing and comparing pairwise distances at each FPS iteration. This leads to significant latency savings while maintaining sampling quality comparable to that of the original FPS. Furthermore, predicting the distance curve enables the sampling process to be decoupled from distance computation. This decoupling exposes new opportunities for parallelizing distance calculations in the proposed sampling, further enhancing efficiency. This approach not only accelerates FPS itself but also benefits subsequent operations such as neighbor search by leveraging pre-computed distances.

\begin{figure*}[t]
\centering
\includegraphics[page=2,width=\textwidth]{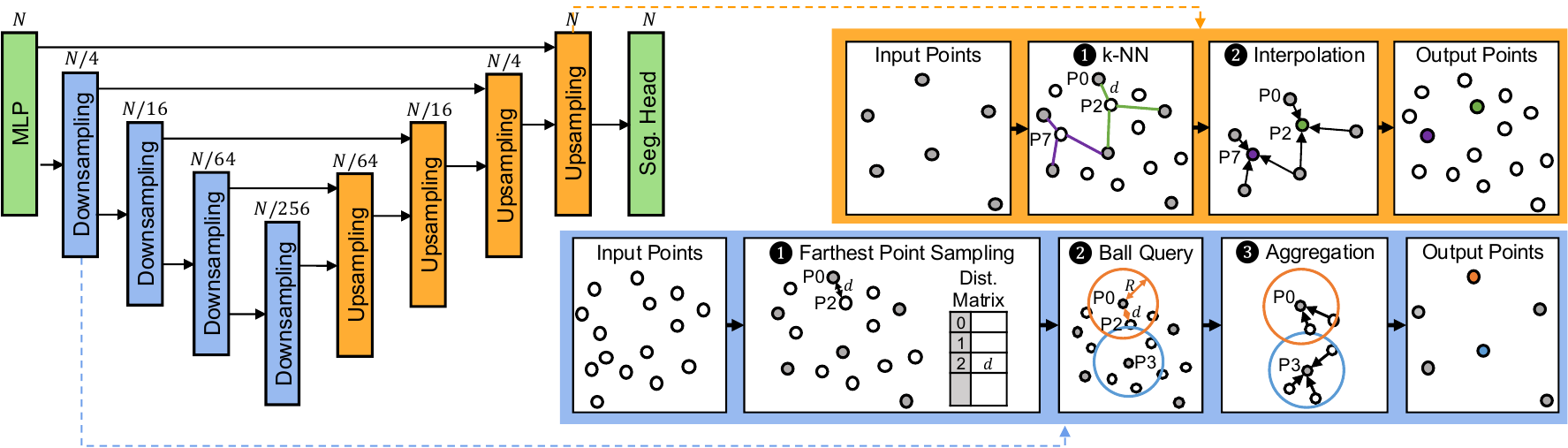}
\caption{Overview of PointNet++ Based Model Architecture}
\label{fig:modelarch}
\vspace{-0.1in}
\end{figure*}

Our key contributions are summarized as follows:

\begin{itemize}
\item We identify two key latency bottlenecks in point cloud models: FPS and neighbor search. Limited parallelism in FPS and redundant distance computations across both operations are the primary sources of inefficiency.

\item We empirically analyze 3D point cloud models and identify the decreasing trend and early structure capture trend.

\item Based on the observations, we introduce a novel sampling approach with sample point distance prediction. With the estimated distance curve, this approach not only accelerates FPS itself but also benefits subsequent neighbor search, which relies on pre-computed distances.

\item We integrate \ours\space into OpenPoints library and experiment on state-of-the-art PointNet++ based models, achieving geomean \geomeanspeedup\space end-to-end speedup on NVIDIA RTX 3090 GPU while preserving accuracy.
\end{itemize}

\section{Background and Related Work}
\label{sec:background}

\subsection{Deep Learning on 3D Point Clouds}

\paragraph{Deep Neural Networks on Raw Point Cloud}
Since the release of PointNet~\cite{pointnet}, which is the first to apply deep neural networks directly to the raw point clouds, numerous successors~\cite{pointnet++, pointweb, assanet, densepoint, dgcnn, fpointnet, randlanet, pointconv, paconv, pointnext, pointvector, pointmetabase} have been developed, improving both model performance and efficiency. PointNet++~\cite{pointnet++} introduces a hierarchical structure to PointNet through sampling and grouping, which has become the core architecture of subsequent PointNet++ based models. PointNeXt~\cite{pointnext} revisits PointNet++, achieving substantial improvements in model performance, while PointVector~\cite{pointvector} and PointMetaBase~\cite{pointmetabase} push the model performance and efficiency even further. Although the architectural details of each model differ, all models follow the PointNet++ structure, utilizing Set Abstraction and Feature Propagation layers as core components. This work focuses on identifying and accelerating the primary latency bottlenecks in PointNet++ based models.

\paragraph{Deep Neural Networks on Voxelized Point Cloud}
Several works~\cite{minkowski, mix3d, cylinder3d, imvoxelnet, occuseg, largekernel3d} propose to voxelize point clouds, using voxels as input rather than raw points. The advent of 3D sparse convolution~\cite{minkowski} has made this voxel-based approach increasingly popular due to its efficiency and high performance. Recently, transformer-based models~\cite{votr, sst, swin3d, swin3d++} operating on voxelized point clouds are gaining significant attention. Although voxelization alleviates the computational cost of mapping operations (i.e., downsampling and neighbor search), it has a notable limitation due to the loss of position information during the voxelization process.

\subsection{Related Work}
\label{related_works}
Numerous solutions have been proposed to accelerate sampling and neighbor search operations in PointNet++ based models. We categorize them into hardware- and purely software-based approaches.
\paragraph{Hardware-Assisted Acceleration}
QuickFPS~\cite{quickfps} introduces a k-d tree based FPS algorithm which reduces computation and memory access in each FPS iteration, along with specialized hardware designed for this method. MARS~\cite{mars} and PTrAcc~\cite{dynamicfps} both propose hardware accelerators that employ a distance filtering technique which skips unnecessary distance computations in each iteration of FPS. Although these techniques achieve substantial speedups in FPS without any loss in sampling quality, they require specialized hardware to fully harness their algorithms, limiting their effectiveness on commodity hardware like GPU. For more thorough evaluation, we compare the speedup of our proposal with pure software version~\cite{quickfps-sw} of QuickFPS in Section~\ref{speedup}, evaluating its effectiveness on GPU.

\paragraph{Software-Only Acceleration}
To mitigate the high latency associated with FPS, various alternative sampling methods have been explored. RandLA-Net~\cite{randlanet} replaces FPS with random sampling and introduces a local feature aggregation module that compensates for the low sampling quality of random sampling. Grid sampling, the most commonly used alternative, is adopted by Grid-GCN~\cite{gridgcn} and KPConv~\cite{kpconv}, offering faster processing with reasonable sampling quality compared to random sampling. Despite the latency advantages of these methods, none have consistently outperformed FPS for overall performance. This is illustrated by the fact that most PointNet++-based models~\cite{pointnet++, assanet, fpointnet, pointnext, pointvector, pointmetabase} continue to rely on FPS, highlighting its broad applicability and superior sampling quality. The experimental results in Appendix B.1 corroborate this claim.

EdgePC~\cite{edgepc} structures the point cloud with morton codes to accelerate FPS and neighbor search on edge GPUs, while Adjustable FPS~\cite{adjustable_fps} accelerates FPS by leveraging the intrinsic locality of the point cloud data. However, both methods demonstrate significant accuracy loss compared to the baseline (i.e., up to 2\% mIoU loss) due to their aggressive approximations, limiting their effectiveness. Prior work~\cite{frugal} accelerates the training of PointNet++ based models by precomputing the minimum point spacing of farthest-sampled points before training and reusing this value across epochs for fast sampling. While this approach effectively speeds up training, it is not suitable for inference, as the minimum spacing value cannot be determined in advance in inference scenarios. Detailed comparisons with our work are provided in Appendix A.4.
\section{Challenges of Point Cloud Models}
\label{sec:analysis}

\begin{figure}[t]
\centering
\includegraphics[width=\columnwidth]{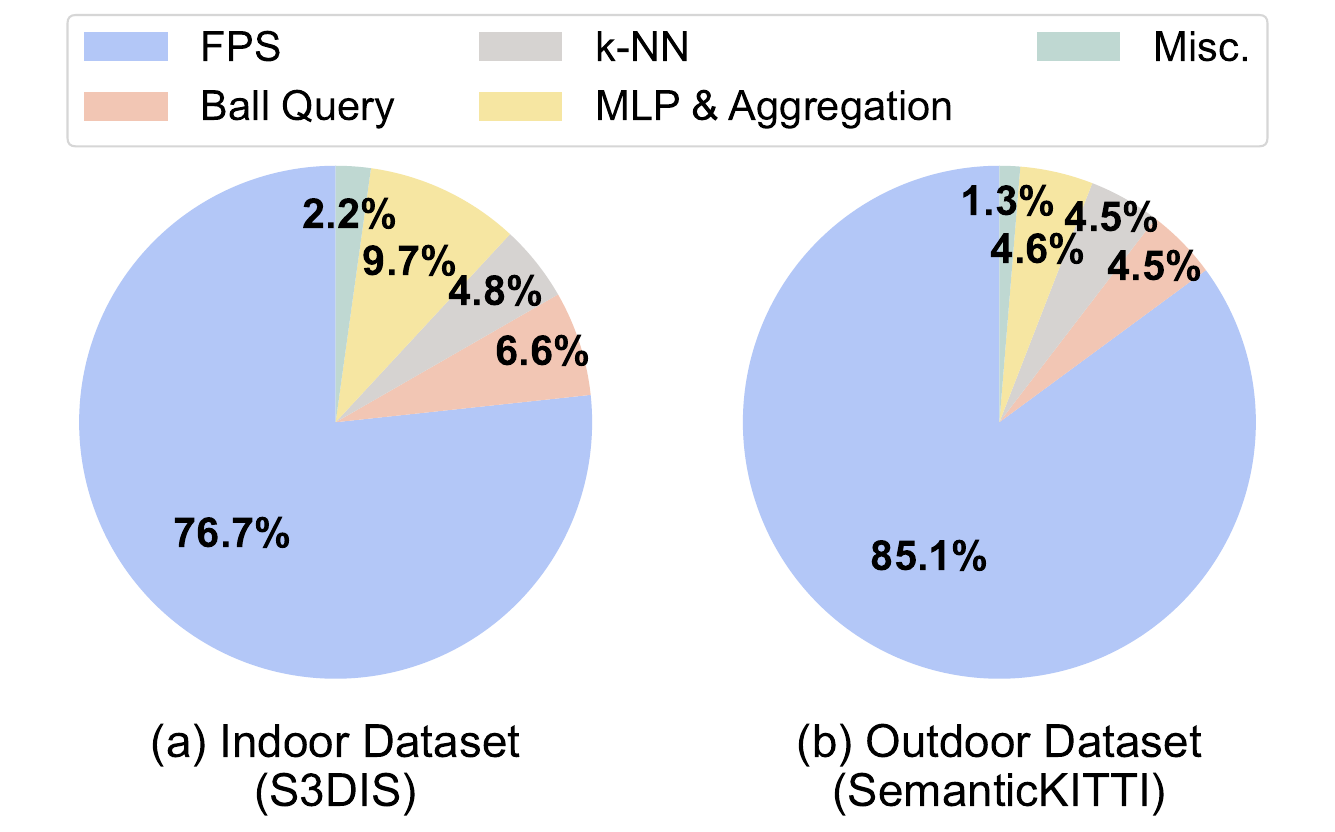}
\caption{Latency Breakdown of PointNet++ Based Models}
\vspace{-0.1in}
\label{fig:breakdown}
\end{figure}

\subsection{Model Architecture Overview}
\label{model_arch}
Figure~\ref{fig:modelarch} illustrates the detailed process of PointNet++-based models. Similar to 2D convolutional neural networks, these models consist of both downsampling and upsampling layers. Commonly referred to as the set abstraction layer, the downsampling layer in PointNet++-based models comprises three main stages, as outlined below:
\renewcommand{\labelitemi}{~}
\begin{itemize}
\item \circled{1} \textbf{Farthest Point Sampling (FPS):} Downsampling begins by selecting $N/stride$ output points from $N$ input points, where $stride$ represents the downsampling rate. Typically, the FPS is used to maintain the shape and structure of the input point cloud.

\item \circled{2} \textbf{Ball Query:} 
Next, a ball query with radius $R$ is performed around each sampled point to gather neighboring points for feature aggregation. 

\item \circled{3} \textbf{Aggregation:} The features of neighboring points identified in the ball query are then processed. A multi-layer perceptron (MLP) is applied to these features, followed by max pooling to aggregate the information for each centroid. The details of this process, e.g., positional embedding~\cite{pointmetabase}, can vary depending on the model architectures.\\
\end{itemize}

The upsampling layer, often referred to as a feature propagation layer, takes the downsampled points from the corresponding downsampling layer as inputs and propagates the features back to the original point cloud. This process is done by following steps.
\begin{itemize}
\item \circled{1} \textbf{k-NN:} For each upsampled point, k-NN is applied to identify its k nearest neighbors among the input points.
\item \circled{2} \textbf{Interpolation:} Based on the k-NN results, neighboring input points are interpolated, concatenated with skip-connected features from the corresponding downsampling layer, and the information is passed through an MLP to propagate the features to the subsequent layer.
\end{itemize}

\subsection{Latency Breakdown}
Figure~\ref{fig:breakdown} presents the latency breakdown of PointMetaBase-L model evaluated on both indoor (S3DIS) and outdoor (SemanticKITTI) scenes. The results show that coordinate-based operations (i.e., FPS, k-NN, and Ball Query) dominate the execution time, consuming significantly more time than feature-based operations (i.e., Aggregation and MLP). These coordinate-based operations account for approximately 88.1\% of the total inference time for indoor scenes and 94.1\% for the outdoor scenes. Among these, FPS emerges as the primary performance bottleneck, requiring the most execution time. Additionally, neighbor search operations, including k-NN and Ball Query, contribute significantly to the overall latency. To address these challenges, we first analyze the root causes of these inefficiencies and propose novel optimization techniques in subsequent sections.

\begin{algorithm}[t]
\caption{Farthest Point Sampling}
\small
\label{alg:fps_algo}
\textbf{Input} $P$: input point cloud, $n$: number of points to sample,\\ $N$: number of original points\\
\textbf{Output} sampled\_idx: indices of sampled points
\begin{algorithmic}[1]
\State sampled\_idx[0] $\gets$ $seed\_idx$ \hspace{2mm} \textcolor{magenta}{// Starts with a random point}
\State min\_dists[$i$] $\gets$ $\infty$ for $i = 0,...,N-1$
\For{$i \gets 1$ to $n - 1$} \hspace{2mm} \textcolor{magenta}{// Not parallelized}
    \For{$j \gets 0$ to $N-1$} \hspace{2mm} \textcolor{magenta}{// Parallelized}
        \State $d_{new}$ $\gets$ \texttt{dist}($P$[sampled\_idx[$i-1$]], $P$[$j$])
        \If {min\_dists[$j$] $>$ $d_{new}$}
            \State min\_dists[$j$] $\gets$ $d_{new}$
        \EndIf
    \EndFor
    \State sampled\_idx[$i$] $\gets$ \texttt{argmax}(min\_dists)
\EndFor
\end{algorithmic}
\end{algorithm}

\subsection{Inefficiencies in Farthest Point Sampling}
\label{inefficiency_fps}
Farthest point sampling operates iteratively by selecting one point at a time. In each iteration, the sampling algorithm identifies the point farthest from the set of the previously sampled points. The detailed process is outlined in Algorithm~\ref{alg:fps_algo}. Initially, a random point in $P$ is chosen as the seed point (Line 1) and added to the \emph{sampled\_idx}. Then, a distance matrix (i.e., \emph{min\_dists}) is created to store the distance between each point in $P$ and the set of the previously sampled points, where the distance from a point to a point set is defined as minimum distance between the point and any point in the set. This matrix is initialized to $\infty$ (Line 2). 
In each iteration, the distance between the most recently sampled point and every other point in $P$ is calculated (Lines 3-7). If a newly calculated distance is smaller than the corresponding entry in the \emph{min\_dists} matrix, the matrix is updated with the smaller value. Finally, the index of the point with the maximum distance in the \emph{min\_dists} is added to the \emph{sampled\_idx} (Line 8). This process is repeated until the specified number of points ($n$) is sampled.

The sequential nature of the FPS algorithm is the primary reason of the inefficiency. Because only one point is sampled per iteration, the $O(N)$ distance calculations required in each iteration must be sequentially executed $n$ times. This tight coupling between sampling and distance calculation significantly limits the potential for parallelism in distance computations.
 
To unlock this parallelism potential, we introduce \emph{Minimum Distance Prediction Sampling (MDPS)} in Section~\ref{flashfps}, a novel sampling strategy that fully decouples sampling from distance computation, enabling greater parallelism and faster processing speeds.

\subsection{Inefficiencies in Neighbor Search}
As explained in Section~\ref{model_arch}, PointNet++ based models employ two types of neighbor search algorithms: Ball Query and k-NN. These operations play a critical role in capturing local geometric relationships in the point cloud. Ball Query identifies neighbor points within a fixed radius around the downsampled points.
k-NN identifies k nearest neighbors among the downsampled points for each original point.

The inefficiencies in these neighbor search operations stem from redundant distance computations in both the FPS and subsequent neighbor search operations. For instance, as shown in Figure~\ref{fig:modelarch}, the distance between P0 and P2 is computed while updating P2’s minimum distance in FPS. However, the same distance is redundantly recalculated in both Ball Query and k-NN. Section~\ref{redundancy_free_ns} introduces \emph{redundancy-free neighbor search} techniques to address this issue.

\section{\ours: Fast Point Cloud Inference}
\label{sec:proposal}

\paragraph{Trends in Minimum Distance of Sample Points}
The FPS iteratively selects the point farthest from the current set of sampled points, which leads to predictable trends in the distance between the sampled points.

\begin{figure}[t]
     \centering
     \begin{subfigure}[b]{0.16\textwidth}
        \centering
        \noindent\includegraphics[width=\textwidth]{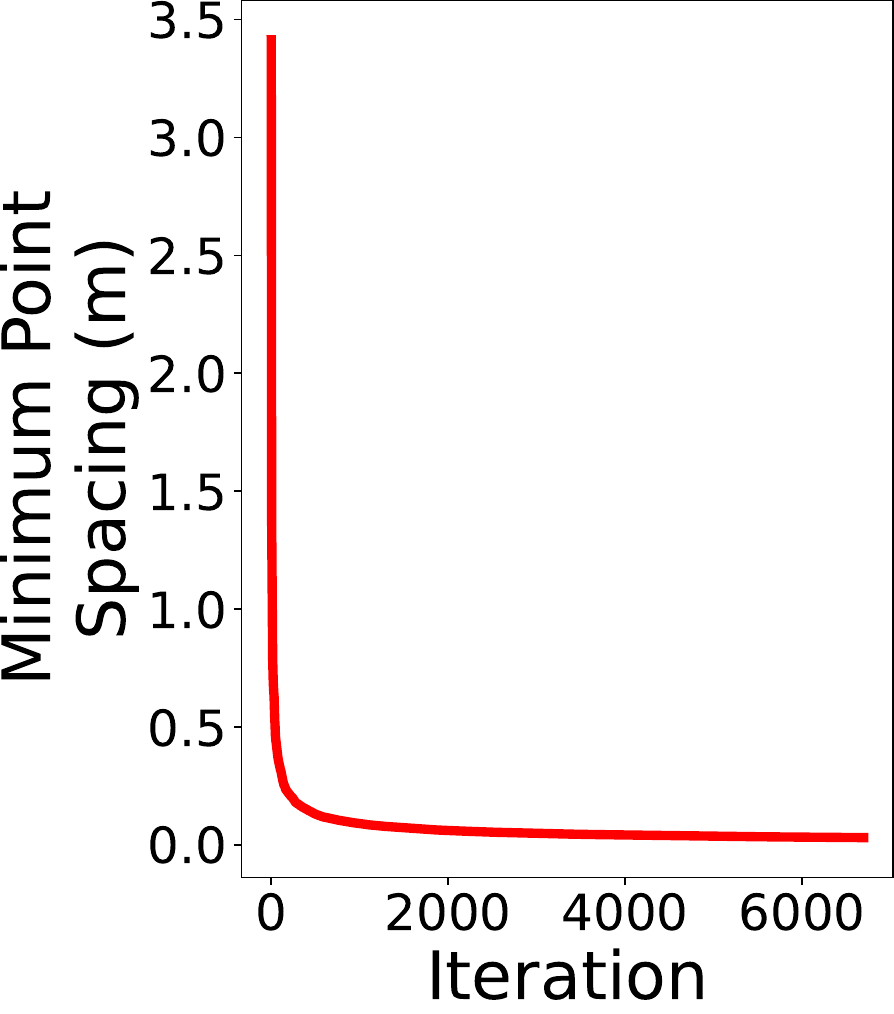} 
        \caption{Minimum Distance Curve of ScanNet scene}
        \label{fig:mps}
     \end{subfigure}
     \hfill
     \begin{subfigure}[b]{0.31\textwidth}
        \centering
        \noindent\includegraphics[page=8,width=\textwidth]{figures/figures-crop.pdf}
        \caption{Visualization of ScanNet scene sampled at 1/10 of the original iterations}
        \label{fig:visualization}
     \end{subfigure}
     \caption{Motivation of Minimum Distance Curve Estimation}
     \vspace{-0.1in}
\end{figure}

\begin{enumerate}
    \item \textit{Decreasing minimum distance}: As sampling proceeds, the pool of the remaining points shrinks, the FPS algorithm selects points closer to those already sampled over time. Thus, the maximized minimum distance (i.e., \texttt{max}(\emph{min\_dists}) in Algorithm~\ref{alg:fps_algo}) decreases, which we define as \emph{minimum distance} for brevity. This is particularly evident in Figure~\ref{fig:mps}, showing a smooth, decreasing curve for various input point clouds.
    \item \textit{Early structure capture}: Importantly, we observe that a few initial sample points are sufficient to capture the overall shape and boundaries of the input point cloud (Figure~\ref{fig:visualization}). This is because the initial sample points tend to be the extremities of the point cloud. 
\end{enumerate}

\paragraph{Optimization Opportunities}
The trend of \textit{early structure capture} suggests that the initial iterations of FPS are important in accurately representing the point cloud's structure. The \textit{Decreasing trend} suggests that the later portion of the minimum distance curve becomes more predictable as sampling progresses. Building upon these trends, if we could accurately estimate this curve using only a few initial FPS iterations, we could maintain a comparable sampling quality while achieving substantial latency reduction by selecting points guided by the predicted curve. 

\vspace{0.2cm}
\noindent\textbf{Proposed Techniques}
We introduce two optimizations by leveraging these opportunities: Minimum Distance Prediction Sampling and Redundancy-Free Neighbor Search. Minimum Distance Prediction Sampling (\oursampling) is a novel sampling strategy designed to approximate the sampling quality of FPS while significantly reducing latency. This approach estimates the distance curve using only a few initial non-parallelizable FPS iterations. Subsequent sample points are selected based on the predicted distances, eliminating the need for pairwise distance comparisons in later iterations.
Furthermore, \oursampling\space enables Redundancy-Free Neighbor Search by reusing precomputed distance information, thereby eliminating redundant computations during subsequent Ball Query and k-NN operations. This optimization further enhances the overall efficiency of PointNet++-based models during inference.

\subsection{Minimum Distance Prediction Sampling}
\label{flashfps}

\begin{figure*}[t]
\centering
\includegraphics[page=4,width=\textwidth]{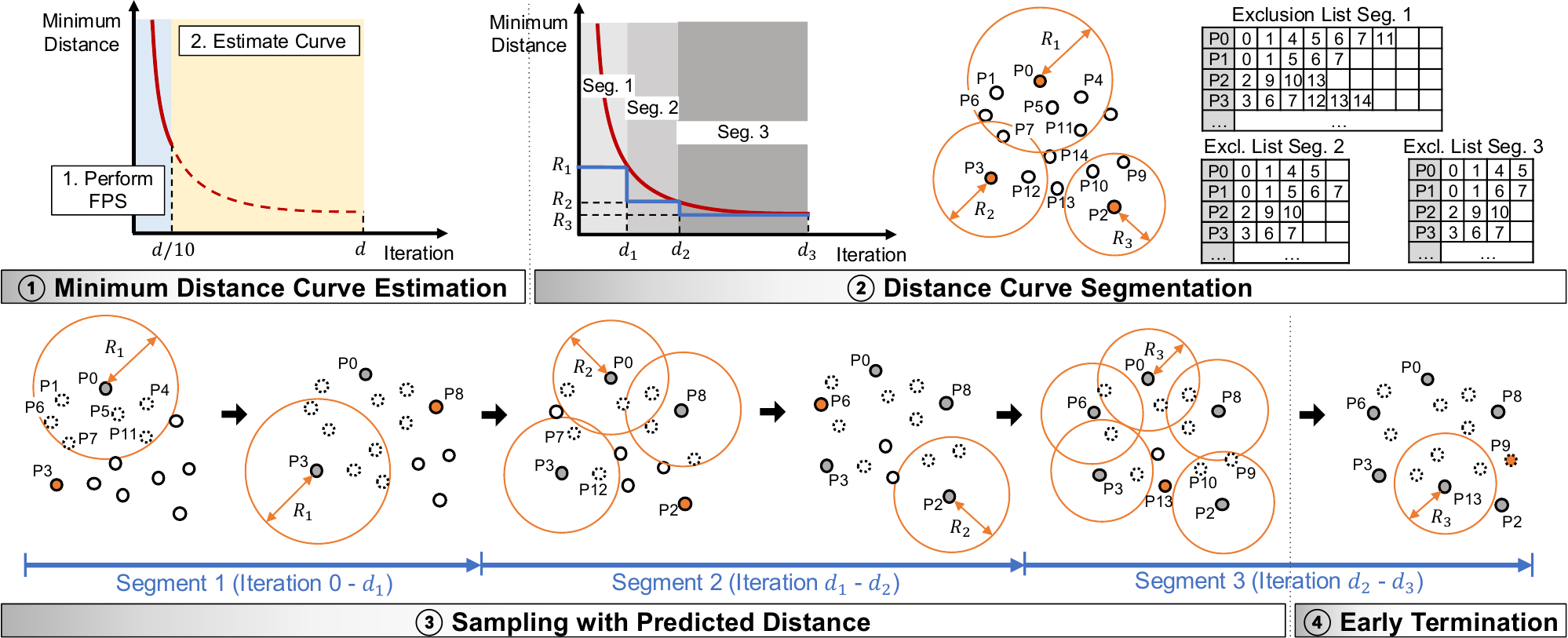}
\caption{Overall Flow of Minimum Distance Prediction Sampling}
\vspace{-0.1in}
\label{fig:flow_1}
\end{figure*}

\textbf{\circled{1} Minimum Distance Curve Estimation}
To build a low-cost estimator for the minimum distance curve (as shown in Figure~\ref{fig:flow_1}), we aim to estimate the curve with minimal error using the results from as few FPS iterations as possible. We formulate the estimator $\mathcal{E}$ as a function which gets initial $p$ segment (where $p$ is the ratio of initial FPS iterations to total iterations, $0 < p < 1$) of the curve $C$ as input and outputs the rest of the estimated curve $\hat{C}$ (i.e., Equation~\ref{eq:estimator}). Note that $C(1)$ represents the minimum distance among the sampled points after all sampling iterations are finished.
\begin{equation}
    \label{eq:estimator}
    \begin{aligned}
        \hat{C}(t) = \mathcal{E}\left(C(t) \mid t \in [0, p]\right) \quad \text{for } t \in (p, 1]
    \end{aligned}
\end{equation}

We have experimented with various models, including polynomial functions and MLPs, and find that MLPs yield the best performance (Appendix A.3). We train a 3-layer MLP on minimum distance curves extracted from the training split of each point cloud dataset, setting $p=0.1$ to balance error and efficiency.
Our estimator achieves a low Mean Absolute Percentage Error (MAPE) of 1.02\%, 0.77\%, and 1.93\% on S3DIS, ScanNet, and SemanticKITTI datasets, respectively, performing better on indoor datasets (S3DIS, ScanNet) due to their more consistent point density and distribution. Appendix B.6 and B.8 discuss ablation study for $p$ and cross-dataset applicability of estimators.

\vspace{0.2cm}
\noindent\textbf{\circled{2} Distance Curve Segmentation}
Ideally, what we want is to sample points whose minimum distance matches the value on the curve at each iteration. However, directly finding the point that yields the exact minimum distance in every iteration is computationally expensive. 

To efficiently sample points along the estimated minimum distance curve, we first divide the curve into multiple segments. Note that each segment is associated with a group of consecutive iterations. Then, the corresponding predicted minimum distances are employed to identify sample point at the segment-level granularity in the sampling process. 

As illustrated in Figure~\ref{fig:flow_1}, the curve is segmented at iterations $d_{1}$, $d_{2}$, and $d_{3}$, resulting in three segments. For each input point, we identify points within a specified radius of each segment boundary (i.e, $R_{1}$, $R_{2}$, and $R_{3}$) and store them in the exclusion list for three segments. This list prevents sampling closely located points by considering the sampled points. 

To optimize computational efficiency, building the exclusion list across all segments is fused into a single GPU kernel. 
By calculating the distance between the points only once, regardless of the number of segments, we minimize the latency overhead. For example, the distance between P0 and P1 is computed only once. If P1 is within $R_{3}$ (i.e., dist(P0, P1) $< R_{3} < R_{2} < R_{1}$), P1 is added to the exclusion list of P0 at all segments. Refer to Table~\ref{table:ablation_level} for latency overhead with an increase in the segment count.
 
Building the exclusion list still requires computing pairwise distances; however, this process is much faster than the original FPS algorithm. This improvement is achieved because all computations are fully parallelizable, as sampling and distance calculations are completely decoupled. For details on the algorithm, refer to Appendix A.1.

\vspace{0.2cm}
\noindent\textbf{\circled{3} Sampling with Predicted Distance}
The sampling process employs the exclusion list. Starting from Segment 1, a seed point (P0 in Figure~\ref{fig:flow_1}) is chosen. In each sampling iteration, points within a radius $R_{1}$ from the most recently sampled point are filtered out based on the exclusion list. We employ bitmaps to record which points are available for sampling. For instance, in the first iteration of Figure~\ref{fig:flow_1}, Point P0, P1, P4, P5, P6, P7, and P11 are excluded based on the exclusion list of Segment 1.  Point P3 is randomly selected among the eight available points for the second iteration in the same segment. Excluded points remain unavailable until the current segment is complete.

\begin{figure}[t]
\centering
\includegraphics[width=\columnwidth]{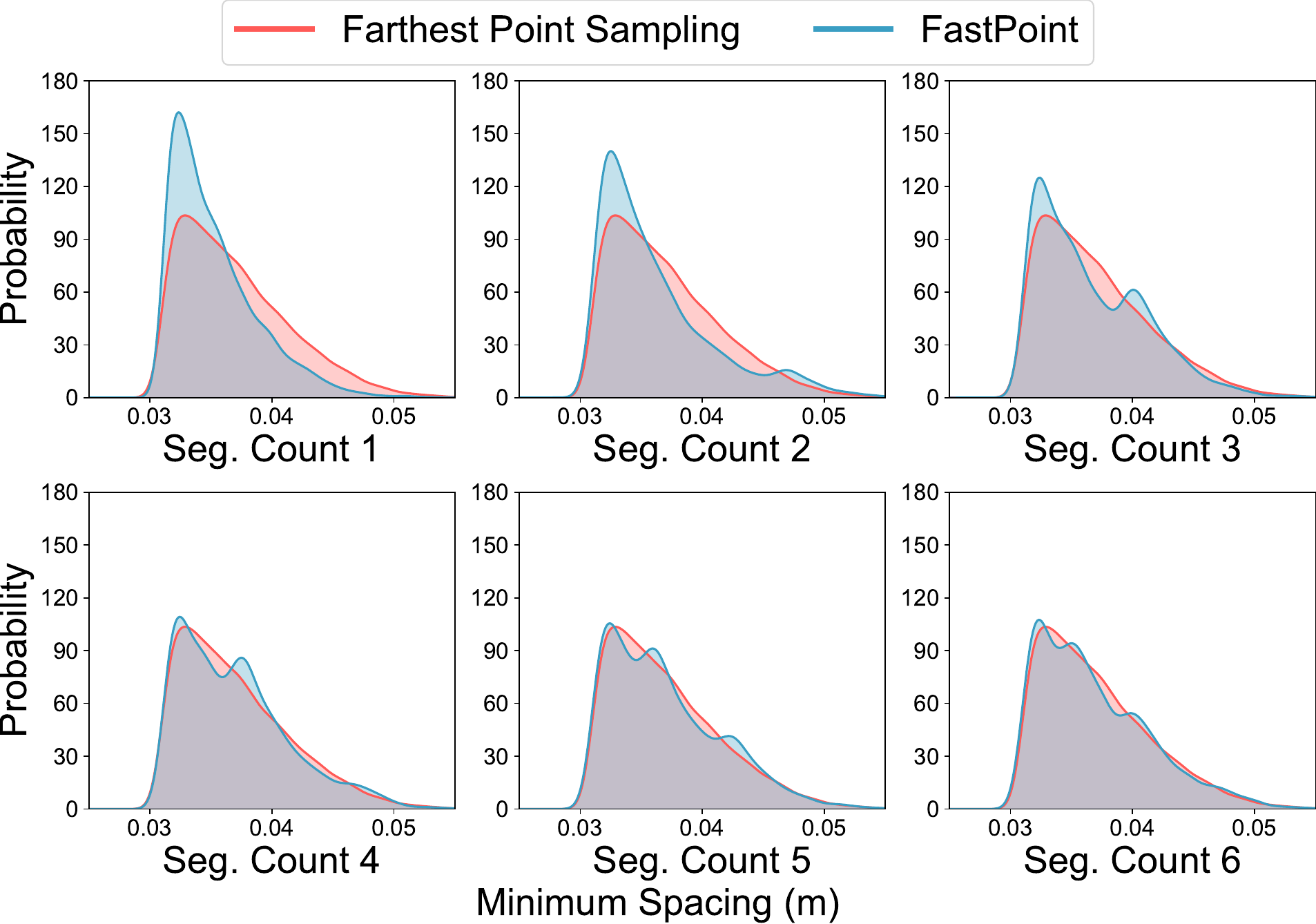}
\caption{Minimum spacing distribution comparison with increasing segment counts.}
\vspace{-0.1in}
\label{fig:converge}
\end{figure}

As the sampling progresses and the segment changes (e.g., from Segment 1 to Segment 2), the radius threshold is adjusted (e.g., from $R_{1}$ to $R_{2}$). This transition changes the set of points that are available for sampling. By iteratively applying this process, we ensure that the distance between sampled points meets or exceeds the threshold for each segment. This results in a minimum distance that lies above the blue step function in the segmented distance curve (i.e., $R_{i}$ serves as the lower bound of the minimum distance in Segment $i$). Increasing the number of segments allows the step function to approach the red minimum distance curve, improving sampling quality and converging towards the quality of FPS. Figure~\ref{fig:converge} demonstrates this convergence. With more segments, the distribution of minimum distance between sampled points becomes closer to that of FPS. Details on implementation are provided in Appendix A.1.

\vspace{0.2cm}
\noindent\textbf{\circled{4} Early Termination}
While the estimated minimum distance curve exhibits low error, overestimation can still negatively impact sampling quality. This is because it can lead to an excessive number of point exclusions, limiting the availability of points in later iterations.
To mitigate this issue, we introduce Early Termination. This technique monitors the availability of points at each iteration. If no points remain, it advances to the next segment earlier than planned. When the final segment is exhausted, Sampling with Predicted Distance is terminated, and the remaining iterations are completed using the original FPS algorithm.
For seamless transition to FPS, we efficiently update the minimum distance matrix ($dists$ in Algorithm 1) by leveraging the exclusion list. This approach avoids recalculating all-to-all distances between points, significantly reducing the search space and minimizing the impact on sampling time. Appendix A.1 describes the details of Early Termination.

\vspace{0.2cm}
\noindent\textbf{Factors Contributing to Speedup}
The primary factors contributing to the latency of \oursampling\space are:
\begin{enumerate}
    \item \emph{Minimum Distance Curve Estimation:} This step, requiring 1/10 of the original FPS iterations, adds only a fraction of the original FPS latency.
    \item \emph{Exclusion List Construction:} The required all-to-all distance calculations are fully parallelizable, resulting in minimal latency overhead. Detailed analysis on this factor is provided in Appendix A.2.
    \item \emph{Sampling Process:} The use of exclusion list eliminates the need for additional distance calculations. Early termination requires additional FPS iterations. However, the impact is minimal: the overhead of 2.02\%, 0.58\%, and 1.18\% of original FPS iterations in S3DIS, ScanNet, and SemanticKITTI dataset, respectively.
\end{enumerate}

Due to these optimizations, \oursampling\space can achieve a 4-5$\times$ speedup relative to the original FPS algorithm. Detailed latency breakdown of \oursampling\space is provided in Appendix B.2.

\begin{figure}[t]
\centering
\includegraphics[page=7,width=\columnwidth]{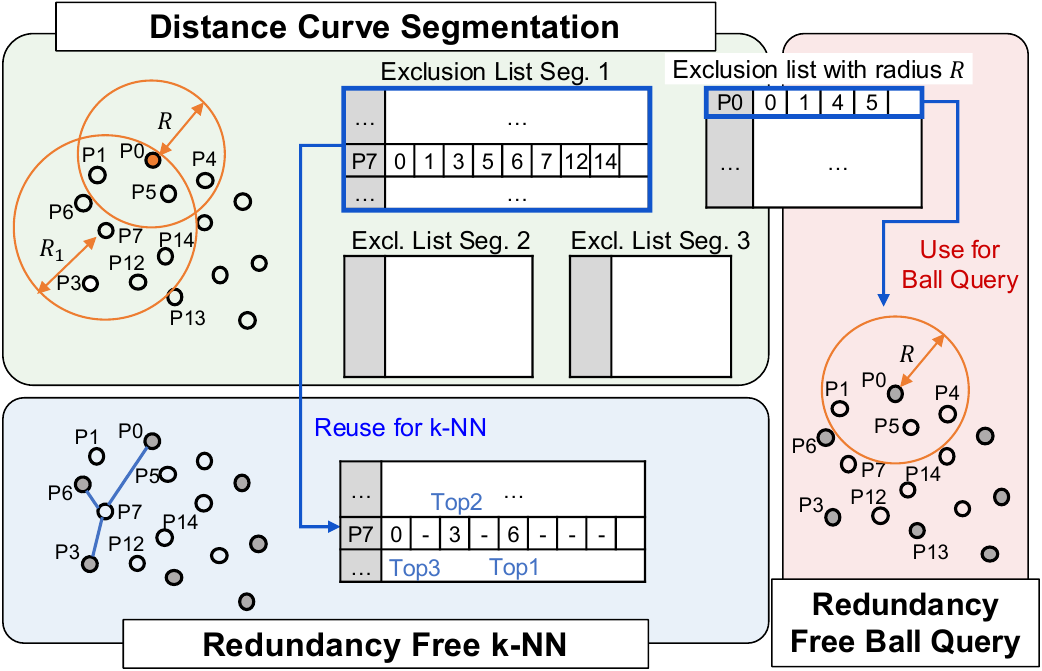}
\caption{Redundancy Free Neighbor Search}
\vspace{-0.1in}
\label{fig:ns}
\end{figure}

\subsection{Redundancy-Free Neighbor Search}
\label{redundancy_free_ns}
The exclusion list in \oursampling\space captures spatial relationships between points, providing an opportunity to optimize subsequent Ball Query and k-NN operations. We introduce Redundancy Free Ball Query and k-NN, which fully leverage the exclusion list from the sampling stage.

\vspace{0.2cm}
\noindent\textbf{Redundancy-Free Ball Query}
Ball query, which identifies neighbors within a radius $R$, is performed after sampling, before aggregation (refer to Figure~\ref{fig:modelarch}). The exclusion list is well-suited for this task, as it contains points within a specified radius of each point. By adding an extra list to the exclusion list for the radius $R$, as illustrated in Figure~\ref{fig:ns}, we can directly use it for ball query. As discussed in Section~\ref{flashfps}, this approach minimizes computational overhead; distances between points are calculated only once, regardless of the number of segments. After sampling, we extract the relevant entries from the corresponding exclusion list for the ball query, which serve as centroids for aggregation.

\vspace{0.2cm}
\noindent\textbf{Redundancy-Free k-NN}
While directly reusing the exclusion list for k-NN is challenging, we can leverage it to reduce the search space. By using the exclusion list of Segment 1, we can limit the search space by eliminating points that are too far away to be potential k-NN neighbors. For example, to find the three nearest neighbors of point P7 in Figure~\ref{fig:ns}, we calculate distances between P7 and the points from the corresponding exclusion list to identify the three nearest neighbors. Because k-NN upsampling relies on the downsampled set, points that were not downsampled are excluded from the search space. This approach significantly reduces the number of distance calculations, leading to substantial latency reduction.
\section{Evaluation}
\label{sec:eval}
\subsection{Methodology}
We evaluate \ours\space on PointMetaBase~\cite{pointmetabase} and PointVector~\cite{pointvector}, two recent models based on PointNet++. We evaluate our proposal on both indoor and outdoor datasets to demonstrate its scalability. We use S3DIS~\cite{s3dis} and ScanNet~\cite{scannet} for indoor datasets and use SemanticKITTI~\cite{semantickitti} for outdoor datasets.

The end-to-end latency is measured for processing all scenes in each dataset's validation split. 
We use mean Intersection over Union (mIoU) as a model accuracy evaluation metric. Detailed evaluation methodologies follow the same approach in PointMetaBase~\cite{pointmetabase}, and PointVector~\cite{pointvector}. We apply \ours\space and the other baseline methods only to the first layer of each model as the first layer's FPS and neighbor search dominates the total computation time (\textgreater 90\%).

We implement \ours\space with custom CUDA kernels and integrate them into OpenPoints library.
We also integrate the software implementation~\cite{quickfps-sw} of QuickFPS~\cite{quickfps} open-sourced by the authors to OpenPoints for comparison. The experiments are conducted using an NVIDIA Geforce RTX 3090 GPU with 24GB of memory. We use CUDA 11.8 and PyTorch 1.10.1 for software setup.

\begin{table}[t]
\setlength{\tabcolsep}{5pt}
\centering
{
    \footnotesize
    \begin{tabulary}{1.0\columnwidth}{C|C|C|C|C}
    \multirow{3}{*}{Dataset} & \multicolumn{4}{c}{Average Minimum Distance (Ratio to Baseline)} \\
    \cline{2-5}
    & \multirow{2}{*}{\shortstack[c]{Baseline\\(FPS)}} & \multirow{2}{*}{\shortstack[c]{Random\\Sampling}} & \multirow{2}{*}{\shortstack[c]{Grid\\Sampling}} & \multirow{2}{*}{\shortstack[c]{\oursampling\\(Ours)}} \\
    & & & & \\
    \hline
    \hline
    \multirow{2}{*}{S3DIS} & \multirow{2}{*}{\shortstack[c]{0.06395}} & \multirow{2}{*}{\shortstack[c]{0.03478\\(54.38\%)}} & \multirow{2}{*}{\shortstack[c]{0.04971\\(77.72\%)}} & \multirow{2}{*}{\shortstack[c]{0.06359\\(99.45\%)}} \\ 
    & & & & \\
    \hline
    \multirow{2}{*}{ScanNet} & \multirow{2}{*}{\shortstack[c]{0.03701}} & \multirow{2}{*}{\shortstack[c]{0.02035\\(54.99\%)}} & \multirow{2}{*}{\shortstack[c]{0.02400\\(64.84\%)}} & \multirow{2}{*}{\shortstack[c]{0.03677\\(99.35\%)}} \\ 
    & & & & \\
    \hline
    \multirow{2}{*}{SemanticKITTI} & \multirow{2}{*}{\shortstack[c]{0.25505}} & \multirow{2}{*}{\shortstack[c]{0.12775\\(50.09\%)}} & \multirow{2}{*}{\shortstack[c]{0.21519\\(84.37\%)}} & \multirow{2}{*}{\shortstack[c]{0.25087\\(98.36\%)}} \\ 
    & & & & \\
    \hline
    \end{tabulary}
}
\vspace{-0.05in}
\caption{Comparison of Sampling Quality among FPS, Random Sampling, Grid Sampling and \oursampling.}
\label{table:sampling_quality}
\vspace{-0.15in}
\end{table}

\subsection{Sampling Quality of \oursampling}
To evaluate sampling quality, we compute the average minimum distance between neighboring sampled points. A larger average distance indicates better coverage of sample points from the perspectives of point distribution and structure preservation. 
As shown in Table~\ref{table:sampling_quality}, \oursampling\space achieves sampling quality comparable to FPS, with an average minimum distance reaching over 99\% of the FPS baseline for S3DIS and ScanNet, and over 98\% for SemanticKITTI. In contrast, Random Sampling and Grid Sampling~\cite{kpconv} exhibit significantly lower quality, with their average minimum distances deviating substantially from the FPS baseline. We also compare the minimum spacing distribution of each sampling method via visualization in Appendix B.9 and evaluate the robustness of \oursampling\space in Appendix B.7.

\subsection{Accuracy Impact of \oursampling}
\label{accuracy}

\begin{table}[t]
\setlength{\tabcolsep}{5pt}
\centering
{
    \footnotesize
    \begin{tabulary}{1.0\columnwidth}{C|C|C|C|C|C}
    & & \multicolumn{4}{c}{mIoU (Diff. to Baseline FPS)} \\ \cline{3-6}
    Model & Dataset & \multirow{2}{*}{\shortstack[c]{Baseline\\(FPS)}} & \multirow{2}{*}{\shortstack[c]{Random\\Sampling}} & \multirow{2}{*}{\shortstack[c]{Grid\\Sampling}} & \multirow{2}{*}{\shortstack[c]{\oursampling\\(Ours)}} \\ 
    & & & & & \\
    \hline
    \hline
    \multirow{6}{*}{PV-L} & \multirow{2}{*}{S3DIS} & \multirow{2}{*}{\shortstack[c]{71.33}} & \multirow{2}{*}{\shortstack[c]{64.83\\(-6.5)}} & \multirow{2}{*}{\shortstack[c]{70.97\\(-0.36)}} & \multirow{2}{*}{\shortstack[c]{71.37\\(+0.04)}} \\
    & & & & & \\ \cline{2-6}
    & \multirow{2}{*}{ScanNet} & \multirow{2}{*}{\shortstack[c]{70.70}} & \multirow{2}{*}{\shortstack[c]{59.94\\(-10.76)}} & \multirow{2}{*}{\shortstack[c]{69.61\\(-1.09)}} & \multirow{2}{*}{\shortstack[c]{70.63\\(-0.07)}} \\
    & & & & & \\ \cline{2-6}
    & \multirow{2}{*}{\shortstack[c]{Semantic\\KITTI}} & \multirow{2}{*}{\shortstack[c]{50.91}} & \multirow{2}{*}{\shortstack[c]{38.46\\(-12.45)}} & \multirow{2}{*}{\shortstack[c]{50.62\\(-0.29)}} & \multirow{2}{*}{\shortstack[c]{50.79\\(-0.12)}} \\ 
    & & & & & \\ \hline
    \multirow{6}{*}{PMB-L} & \multirow{2}{*}{S3DIS} & \multirow{2}{*}{\shortstack[c]{69.72}} & \multirow{2}{*}{\shortstack[c]{68.21\\(-1.51)}} & \multirow{2}{*}{\shortstack[c]{69.54\\(-0.18)}} & \multirow{2}{*}{\shortstack[c]{69.74\\(+0.02)}} \\ 
    & & & & & \\ \cline{2-6}
    & \multirow{2}{*}{ScanNet} & \multirow{2}{*}{\shortstack[c]{70.86}} & \multirow{2}{*}{\shortstack[c]{67.66\\(-3.2)}} & \multirow{2}{*}{\shortstack[c]{70.36\\(-0.50)}} & \multirow{2}{*}{\shortstack[c]{70.89\\(+0.03)}} \\
    & & & & & \\ \cline{2-6}
    & \multirow{2}{*}{\shortstack[c]{Semantic\\KITTI}} & \multirow{2}{*}{\shortstack[c]{52.19}} & \multirow{2}{*}{\shortstack[c]{47.26\\(-4.93)}} & \multirow{2}{*}{\shortstack[c]{52.27\\(+0.08)}} & \multirow{2}{*}{\shortstack[c]{52.09\\(-0.10)}} \\ 
    & & & & & \\ \hline
    \end{tabulary}
}
\vspace{-0.05in}
\caption{Accuracy comparison. PV and PMB stand for PointVector and PointMetaBase.}
\label{table:accuracy_impact}
\vspace{-0.15in}
\end{table}

We analyze the impact of \oursampling\space on accuracy by applying it to the inference stage of PointNet++ models initially trained with FPS. Since \oursampling\space is designed to closely replicate the FPS sampling pattern, it is compatible with FPS-trained models, thereby minimizing accuracy degradation. Note that Redundancy-Free Neighbor Search does not introduce any approximation, thus has no impact on accuracy. To compare the effects of different sampling methods, we also evaluate random sampling and grid sampling\footnote[1]{Grid sampling is applied only during inference. Appendix B.1 shows results for cases where both training and inference utilize grid sampling.}.

Table~\ref{table:accuracy_impact} shows that \oursampling\space excels in maintaining accuracy, with negligible differences observed across all datasets and models. This highlights its effectiveness in replicating FPS behavior. However, the grid sampling exhibits larger and less consistent accuracy drops. When applied only during inference, it leads to declines of up to 1.09\% (PointVector-L, ScanNet). While it outperforms \ours\space in a single instance (PointMetaBase-L, SemanticKITTI), its performance is generally inferior, with significant accuracy reductions in other scenarios. The random sampling shows considerable accuracy drops across all data points.

Overall, \ours\space effectively preserves accuracy by closely approximating the sampling pattern of FPS, achieving accuracy levels near-identical to FPS without retraining. Latency-accuracy comparisons (Appendix B.5) show that \ours\space resides above the pareto front, emphasizing the effectiveness of \ours.

\begin{table}[t]
\setlength{\tabcolsep}{5pt}
\centering
{
    \footnotesize
    \begin{tabulary}{1.0\columnwidth}{C|C|C|C|C}
    & \multirow{3}{*}{\shortstack[c]{Number of\\Segments}} & \multicolumn{3}{c}{Evaluation Metrics} \\ \cline{3-5}
    Method & & \multirow{2}{*}{\shortstack[c]{Accuracy\\(mIoU)}} & \multirow{2}{*}{\shortstack[c]{Sampling \\Time (ms)}} & \multirow{2}{*}{\shortstack[c]{Sampling\\Quality}} \\ 
    & & & & \\
    \hline
    \hline
    FPS & - & 70.70 & 85.83 & 0.03701 \\ \hline
    \multirow{7}{*}{\shortstack[c]{\oursampling\\(Ours)}} 
    & 1 Segment & 70.34 & 18.62 & 0.03532 \\ 
    & 2 Segments & 70.60 & 19.43 & 0.03598 \\ 
    & 3 Segments & 70.46 & 19.79 & 0.03636 \\ 
    & 4 Segments & 70.54 & 21.01 & 0.03662 \\ 
    & 5 Segments & 70.52 & 21.41 & 0.03665 \\ 
    & 6 Segments & 70.63 & 22.06 & 0.03662 \\ 
    & 7 Segments & 70.63 & 22.75 & 0.03664 \\ 
    \hline
    \end{tabulary}
}
\caption{Comparison of mIoU, sampling time, and sampling quality for FPS and \oursampling\space across segments 1 to 7. Sampling time is measured with a single scene of ScanNet dataset. PointVector-L model is used for mIoU comparison.}
\label{table:ablation_level}
\vspace{-0.1in}
\end{table}

\subsection{Latency Reduction of \ours}
\label{speedup}

\begin{figure}[t]
    \centering
    \includegraphics[width=1\linewidth]{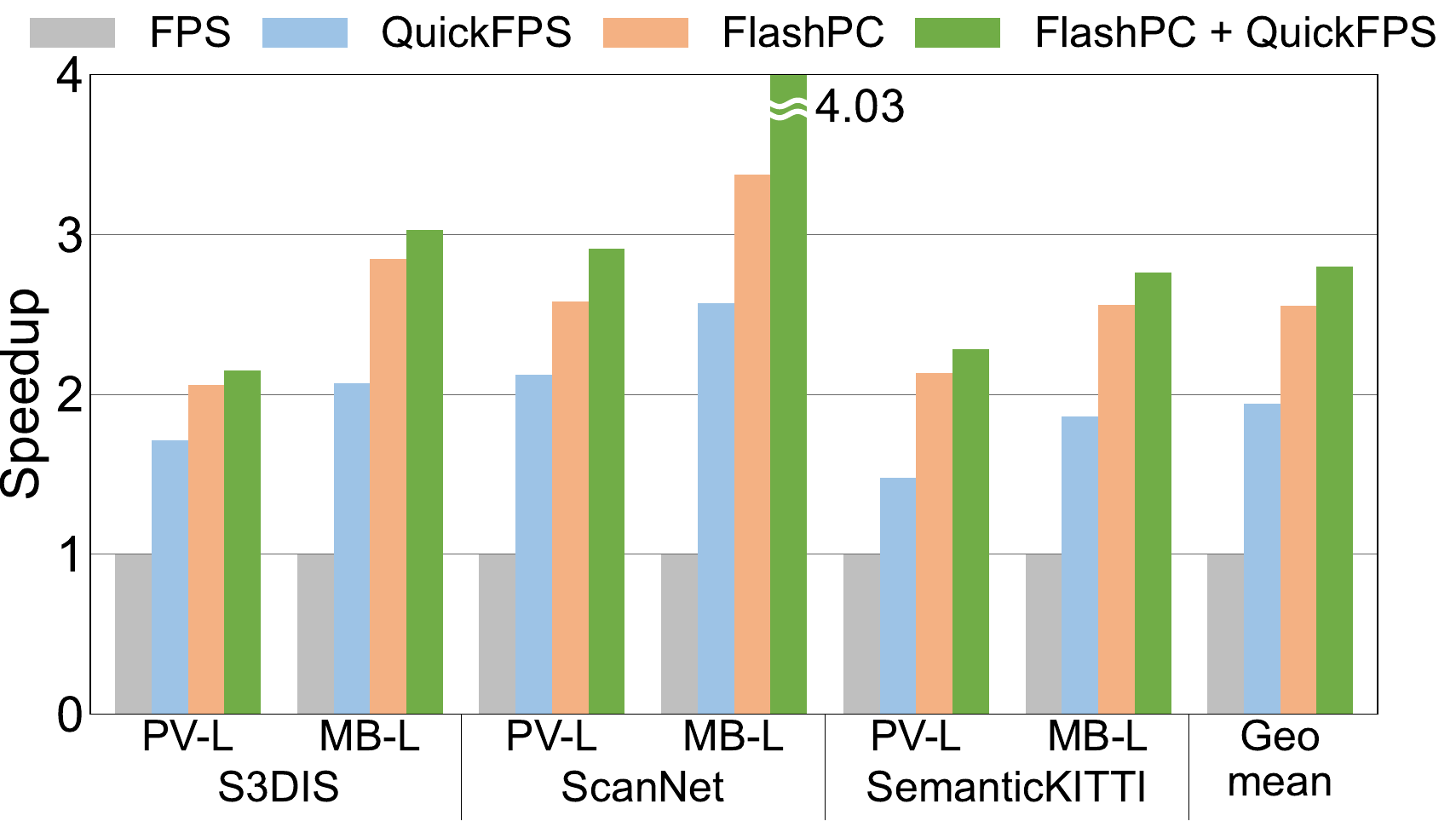}
    \caption{End-to-end speedup of \ours\space and QuickFPS.}
    \label{fig:e2e_speedup}
    \vspace{-0.1in}
\end{figure}

To evaluate the impact of \ours\space on end-to-end latency, we compare its latency to baseline FPS and QuickFPS. Note that FastPoint includes both \oursampling\space and Redundancy-Free Neighbor Search. As explained in Section~\ref{related_works}, QuickFPS accelerates FPS using k-d tree without approximation. Since \oursampling\space also contains FPS steps when predicting the minimum distance curve, we can achieve further latency reductions by replacing them with QuickFPS, meaning that \ours\space and QuickFPS can be synergetic. 

Figure~\ref{fig:e2e_speedup} presents the results of this comparison.
Across S3DIS, ScanNet, and SemanticKITTI datasets, \ours\space achieves substantial latency reductions, with a geomean end-to-end speedup of \geomeanspeedup\space compared to baseline FPS. Integrating QuickFPS with \ours\space further reduces the execution latency achieving a geomean end-to-end speedup of 2.76$\times$ compared to baseline FPS and 1.41$\times$ compared to standalone QuickFPS. Speedup specific to the sampling and neighbor search are provided in Appendix B.3 and B.4. 

These findings substantiate the effectiveness of \ours\space in reducing latency, surpassing existing methods like QuickFPS. Moreover, the successful combination of \ours\space with QuickFPS demonstrates its compatibility with existing FPS optimizations. Scalability to Non-PointNet++ based models are discussed in Appendix B.10.

\subsection{Ablation Study}

\noindent\textbf{Segment Count} 
We evaluate the impact of varying the segment count on accuracy, sampling time, and sampling quality (i.e, average minimum distance). 
Table~\ref{table:ablation_level} shows that while sampling time increases gradually with more segments due to the exclusion list construction (refer to Section~\ref{flashfps}), both sampling quality and accuracy improve. However, they get saturated beyond a certain number of segments. Based on the empirical analysis, we use 6 segments by default, where accuracy saturates for all workloads.

\begin{table}[t]
\setlength{\tabcolsep}{5pt}
\centering
{
    \scriptsize
    \begin{subtable}[h]{0.51\columnwidth}
        \begin{tabulary}{1.0\columnwidth}{C|C|C|C}
        Model & Dataset & Method & Speedup \\ 
        \hline
        \hline
        \multirow{6}{*}{PV-L} & \multirow{2}{*}{S3DIS} 
        & \oursampling & 1.90$\times$ \\ 
        & & All & 1.98$\times$ \\ 
        \cline{2-4}
        & \multirow{2}{*}{ScanNet} 
        & \oursampling & 2.33$\times$ \\ 
        & & All & 2.52$\times$ \\ 
        \cline{2-4}
        & \multirow{2}{*}{\shortstack[c]{Sematic\\KITTI}} 
        & \oursampling & 2.07$\times$ \\ 
        & & All & 2.20$\times$ \\ 
        \hline
        \multirow{6}{*}{PMB-L} & \multirow{2}{*}{S3DIS} 
        & \oursampling & 2.50$\times$ \\ 
        & & All & 2.82$\times$ \\ 
        \cline{2-4}
        & \multirow{2}{*}{ScanNet} 
        & \oursampling & 2.89$\times$ \\ 
        & & All & 3.38$\times$ \\ 
        \cline{2-4}
        & \multirow{2}{*}{\shortstack[c]{Sematic\\KITTI}} 
        & \oursampling & 2.31$\times$ \\ 
        & & All & 2.62$\times$ \\ 
        \hline
        \end{tabulary}
        \caption{Comparison of component-wise end-to-end speedup of \ours.}
        \vspace{-0.1in}
        \label{table:ablation_speedup}
    \end{subtable}
    \hfill
    \begin{subtable}[h]{0.477\columnwidth}
        \begin{tabulary}{1.0\columnwidth}{C|C|C}
        Model & \# of Points & Speedup \\ 
        \hline
        \hline
        \multirow{6}{*}{PV-L} 
        & 16,000 & 1.33$\times$ \\ 
        & 32,000 & 2.41$\times$ \\ 
        & 48,000 & 2.68$\times$ \\ 
        & 64,000 & 3.01$\times$ \\ 
        & 80,000 & 3.25$\times$ \\ 
        & 96,000 & 3.40$\times$ \\ 
        \hline
        \multirow{6}{*}{PMB-L} 
        & 16,000 & 1.58$\times$ \\ 
        & 32,000 & 2.64$\times$ \\ 
        & 48,000 & 2.91$\times$ \\ 
        & 64,000 & 3.06$\times$ \\ 
        & 80,000 & 3.30$\times$ \\ 
        & 96,000 & 3.44$\times$ \\ 
        \hline
        \end{tabulary}
        \caption{Comparison of speedup as the number of points increases.}
        \vspace{-0.1in}
        \label{table:point_speedup}
    \end{subtable}
}

\caption{Ablation study for speedup breakdown and scalability.}
\vspace{-0.1in}
\end{table}

\vspace{0.1cm}
\noindent\textbf{Component-wise Speedup}
To understand the individual contributions of each component in \ours, we evaluate the end-to-end speedup by sequentially applying \oursampling\space and Redundancy Free Neighbor Search.  Table~\ref{table:ablation_speedup} shows that \oursampling\space alone provides significant speedup. Incorporating Redundancy Free Neighbor Search further provides  additional improvement, demonstrating that each component makes a distinct contribution to the overall speedup.

\subsection{Scalability Study}
To assess the scalability of \ours, we evaluate the end-to-end speedup compared to FPS by varying point cloud sizes. Table~\ref{table:point_speedup} shows that \ours\space consistently achieves higher speedups as the number of points increases. This trend highlights the efficiency of our approach, particularly when processing larger point clouds where the computational cost of the FPS becomes a significant bottleneck.
\section{Conclusion}
\label{sec:conclusion}
3D point clouds are critical for representing and understanding 3D scenes, and deep learning models like PointNet++ have shown great promise in processing this data. However, computational efficiency remains a key challenge. We propose FastPoint, a software acceleration technique that leverages predictable minimum distances between sampled points to efficiently identify subsequent points without exhaustive computations. Our evaluation shows a 2.55$\times$ speedup over baseline FPS while maintaining accuracy, enabling low-cost deployment of PointNet++ based models.

\section*{Acknowledgements}

This work was supported by the Ministry of Science and ICT (MSIT), Korea, under the following grants: the Global Scholars Invitation Program (RS-2024-00456287), the Graduate School of Artificial Intelligence Semiconductor grant (IITP-2025-RS-2023-00256081), and the Artificial Intelligence Innovation Hub (No.~2021-0-02068), all supervised by the Institute for Information \& Communications Technology Planning \& Evaluation (IITP). Additional support was provided by the National Research Foundation of Korea (NRF) grant (RS-2024-00405857) funded by the Korea government (MSIT), and the Google Faculty Research Award. The source code is available at \url{https://github.com/SNU-ARC/FastPoint.git}. Jae W. Lee and Hongil Yoon are the corresponding authors.

{
    \small
    \bibliographystyle{ieeenat_fullname}
    \bibliography{main}
}

\clearpage

\maketitle
\appendix

\section{Supplementary Materials for MDPS}
\subsection{MDPS Algorithm}
\label{appendix:algorithm}

In this section, we present detailed algorithm (Algorithm~\ref{alg:flashfps}) of Minimum Distance Prediction Sampling (MDPS).
\paragraph{1. Minimum Distance Curve Estimation}
Minimum distance curve estimation starts by performing FPS for 1/10 of the original number of iterations. The key difference between original FPS and this operation (Line 3-11) is that the maximum of minimum distance (i.e., \texttt{max}(dists)) must be saved at each iteration. These values serve as input to the minimum distance curve estimator, which predicts the rest of the curve. The computed and estimated values are then concatenated to form the complete curve (Line 13).

\paragraph{2. Distance Curve Segmentation}
We divide the distance curve into $nseg$ segments and find the points within a specified radius of segment boundaries (i.e., mdc[$n * seg/nseg$]) for each input point. Distance between an input point $P[i]$ and a query point $P[j]$ is calculated (Line 18), and the index of query point is added to the exclusion list if the distance is smaller than the corresponding threshold (Line 19-21). Distance computation between points (Line 18) is performed outside the loop, ensuring that the amount of distance computation remains independent of $nseg$.

\paragraph{3. Sampling with Predicted Distance}
After initializing the bitmap for all segments to 1 (Line 24), sampling begins. Sampling consists of three main stages: bitmap update, sampling, and sampling availability check. First, we check the entry of exclusion list that corresponds to the previously selected point (Line 30) and update the bitmap with value 1 according to the entry (Line 31-32). Note that we update not only the bitmap of current segment, but also the bitmap of subsequent segments. This ensures that the bitmaps for all subsequent segments remain up to date, facilitating smooth transitions to the next segment during sampling. Next, we sample any point that is available by finding the index of a bitmap entry with a value of 1 (Line 33). If no such entry is found, the process moves to the next segment by incrementing the value $seg$ (Line 34-36). When the final segment is exhausted, we terminate the sampling process and proceed to Early Termination stage (Line 26-28).

\begin{algorithm}
\caption{Minimum Distance Prediction Sampling}
\small
\label{alg:flashfps}
\textbf{Input} $P$: input point cloud, $n$: number of points to sample,\\ $N$: number of original points\\
\textbf{Output} sampled\_idx: indices of sampled points
\begin{algorithmic}[1]
\State \textcolor{magenta}{// 1. Minimum Distance Curve Estimation}
\State \textcolor{magenta}{// 1-1. 1/10 FPS iterations}
\State sampled\_idx[0] $\gets$ $seed$
\State dists[$i$] $\gets$ $\infty$ for $i = 0,...,N-1$
\For{$i \gets 1$ to $n/10$}
    \For{$j \gets 0$ to $N-1$}\hspace{2mm} \textcolor{magenta}{// Parallelized}
        \State $d_{new}$ $\gets$ \texttt{dist}($P$[sampled\_idx[$i-1$]], $P$[$j$])
        \If {dists[$j$] $>$ $d_{new}$}
            \State dists[$j$] $\gets$ $d_{new}$
        \EndIf
    \EndFor
    \State sampled\_idx[$i$] $\gets$ \texttt{argmax}(dists)
    \State mdc[$i$] $\gets$ \texttt{max}(dists)
\EndFor
\State \textcolor{magenta}{// 1-2. Estimation}
\State mdc $\gets$ \texttt{concat}(mdc[$0:n/10$], $\mathcal{E}$(mdc[$0:n/10$]))
\State
\State \textcolor{magenta}{// 2. Distance Curve Segmentation}
\For{$i \gets 0$ to $N-1$} \hspace{2mm} \textcolor{magenta}{// Parallelized}
    \For{$j \gets 0$ to $N-1$} \hspace{2mm} \textcolor{magenta}{// Parallelized}
        \State $d$ $\gets$ \texttt{dist}($P$[$i$], $P$[$j$])
        \For{$seg \gets 0$ to $nseg$}
            \If {$d$ $<$ mdc[$n * seg/nseg$]}
                \State excl\_list$_{seg}$[$i$]\texttt{.append}($j$)
            \EndIf
        \EndFor
    \EndFor
\EndFor
\State
\State \textcolor{magenta}{// 3. Sampling with Predicted Distance}
\State bitmap$_{seg}$[$i$] $\gets 1$ for $i = 0,...,N-1$, $seg = 1,...,nseg$
\For{$i \gets 1$ to $n-1$}\hspace{2mm}
    \State $seg$ $\gets$ $\operatorname{max}(\operatorname{div}(i, n/nseg), seg)$
    \If {$seg > nseg$}
        \State last\_idx $\gets$ $i$
        \State \textbf{break}\hspace{2mm}\textcolor{magenta}{// Early Termination}
    \EndIf
    \For{$l \gets seg$ to $nseg$}\hspace{2mm}\textcolor{magenta}{// Parallelized}
        \State list $\gets$ excl\_list$_{l}$[sampled\_idx[$i-1$]]
        \For{$j \gets 0$ to \texttt{len}(list)}\hspace{2mm} \textcolor{magenta}{// Parallelized}
            \State bitmap$_{l}$[list[j]] $\gets$ $0$
        \EndFor
    \EndFor
    \State sampled\_idx[$i$] $\gets$ \texttt{findAnyOne}(bitmap$_{seg}$)
    \If {sampled\_idx[$i$] $== -1$}\hspace{2mm}\textcolor{magenta}{// No point available}
        \State $seg$ $\gets$ $seg + 1$
        \State $i$ $\gets$ $i - 1$
    \EndIf
\EndFor
\State
\State \textcolor{magenta}{// 4. Early Termination}
\State \textcolor{magenta}{// 4-1. Distance Matrix Update}
\For{$i \gets 0$ to $N-1$} \hspace{2mm} \textcolor{magenta}{// Parallelized}
    \For{$j \gets 0$ to \texttt{len}(excl\_list$_{1}[$i$]$)}
        \If {excl\_list$_{1}$[$i$][$j$] in sampled\_idx}
            \State $d_{new}$ $\gets$ \texttt{dist}($P$[$i$], $P$[excl\_list$_{1}$[$i$][$j$]])
            \If {dists[$i$] $>$ $d_{new}$}
                \State dists[$i$] $\gets$ $d_{new}$
            \EndIf
        \EndIf
    \EndFor
\EndFor
\State \textcolor{magenta}{// 4-2. Remainder FPS}
\For{$i \gets \text{last\_idx}$ to $n-1$}
    \For{$j \gets 0$ to $N-1$}\hspace{2mm} \textcolor{magenta}{// Parallelized}
        \State $d_{new}$ $\gets$ \texttt{dist}($P$[sampled\_idx[$i-1$]], $P$[$j$])
        \If {dists[$j$] $>$ $d_{new}$}
            \State dists[$j$] $\gets$ $d_{new}$
        \EndIf
    \EndFor
    \State sampled\_idx[$i$] $\gets$ \texttt{argmax}(dists)
\EndFor
\end{algorithmic}
\end{algorithm}

\paragraph{4. Early Termination}
If the sampling stage terminates before acquiring the desired number of points $n$, we make a transition to Farthest Point Sampling (FPS). For this transition, the FPS distance matrix is initialized with the minimum distances between the input points and the already sampled point set. To optimize this process, instead of computing all-to-all distances, we leverage the exclusion list from the first segment (i.e., excl\_list$_{1}$) to limit the search space when identifying the closest sampled point for each input point (Lines 43–46). Once the distance matrix is updated, the standard FPS algorithm is applied to complete the remaining iterations (Lines 48–53).

\subsection{Analysis on Exclusion List Construction}
\label{appendix:time_complexity}
In this section, we analyze the computational complexity of exclusion list construction and compare it with that of the baseline FPS. The exclusion list construction requires all-to-all distance calculations, resulting in a computational complexity of $O(N^2)$, while FPS has a complexity of $O(Nn)$, where $N$ and $n$ denote the number of input points and sampled points, respectively. 

Despite the higher theoretical complexity, MDPS is significantly faster than FPS due to two key factors: its high degree of parallelism (as discussed in the paper) and its efficient utilization of GPU Streaming Multiprocessors (SMs). 

FPS only utilizes single SM since parallelizing the distance matrix update across SMs requires frequent SM-to-SM communication every iteration, incurring considerable latency overhead. In contrast, exclusion list construction has no such restriction and fully leverages available SMs.

Thus, considering the utilization of SMs, the complexity of MDPS becomes $O(N^2/p)$ (where $p$ represents the number of SMs), while the FPS remains bound by $O(N^2/stride)$ (where $n$=$N/stride$). This translates to considerable speedup considering the large number of SMs in GPUs (81 in RTX 3090) and typically smaller downsampling $stride$ values in point cloud models ($stride \leq 4 $ for all OpenPoints library models).

\subsection{Minimum Distance Curve Estimator}
\label{appendix:estimator}
To develop a minimum distance curve estimator, we explore various models, including power functions and multi-layer-perceptrons (MLPs).

For the power function-based estimator, the equation \( y = \frac{a}{x^n} \) yields the best results. Here, $n$ determines the decreasing speed of the curve while $a$ reflects the density of each input point cloud. The parameter $n$ is determined by fitting the power function to the entire curve obtained from the training split of each dataset. In contrast, $a$ is determined dynamically during inference, using only the first 1/10 segment of the curve. 

For the MLP-based estimator, we use a lightweight three-layer MLP (32-128-128-64). The estimator is trained to take the first 1/10 segment of the curve (i.e., represented by 32 values) as input, and estimate the remaining curve (i.e., represented by 64 values). Training is performed using input-output pairs derived from the minimum distance curves of each training dataset. Training is conducted for 20 epochs on both S3DIS and ScanNet, and 30 epochs on SemanticKITTI. We use a batch size of 1 and the SGD optimizer with a learning rate of 0.01 for all datasets.

\begin{table}[t]
\centering
{
    \small
    \begin{tabulary}{1.0\columnwidth}{C|C|C}
    Method & Dataset & MAPE (\%) \\ 
    \hline
    \hline
    \multirow{3}{*}{MLP} 
    & S3DIS & 1.0194 \\ 
    & ScanNet & 0.7663 \\ 
    & SemanticKITTI & 1.9318 \\ 
    \hline
    \multirow{3}{*}{Power Function} 
    & S3DIS & 1.4746 \\ 
    & ScanNet & 1.8816 \\ 
    & SemanticKITTI & 6.6118 \\ 
    \hline
    \end{tabulary}
}
\caption{Comparison of MAPE values for MLP and Power function based estimators across datasets.}
\label{table:curve_estimator}
\end{table}

For evaluation, we use the validation split of each dataset and measure Mean Absolute Percentage Error (MAPE) between the predicted curve and the real curve. Results in Table~\ref{table:curve_estimator} demonstrate that MLP-based method significantly outperforms power function-based methods, especially on SemanticKITTI dataset. These results highlight the MLP-based method's superior ability to handle the varying point density and distribution of outdoor datasets. Based on these results, we adopt the MLP-based method as our primary approach for minimum distance curve estimation.

\subsection{Comparison with L-FPS}
\label{appendix:frugal}
The L-FPS~\cite{frugal} algorithm is designed to accelerate the FPS process in the training pipeline of PointNet-based models. L-FPS performs FPS once prior to training to find out the minimum distance between the sampled points. L-FPS then filters out points that are closer than this distance threshold to produce the sampling results for each epoch. While this approach effectively speeds up training, it is not suitable for inference, as the minimum distance value cannot be determined in advance in inference scenarios. Additionally, L-FPS relies on a single threshold (i.e., minimum distance at the final iteration of FPS) during filtering, which results in inferior sampling results compared to FPS (i.e., comparable to 1 segment results in Figure 5).

\section{Additional Experiments}

\subsection{Model Performance Comparison with FPS and Grid Sampling}
\label{appendix:grid}

In this section, we train PointNet++ based models with Grid Sampling and compare the model performance with those trained with FPS (i.e., baseline). While Grid Sampling occasionally achieves better results (e.g., PointVector on SemanticKITTI), FPS generally outperforms Grid Sampling in most cases with a significant performance gain. This consistent trend highlights the robustness of FPS in adapting to various datasets and tasks. Given that {\oursampling} closely matches the performance of FPS, it demonstrates a distinct advantage over Grid Sampling in this context.

\begin{table}[t]
\setlength{\tabcolsep}{5pt}
\centering
{
    \small
    \begin{tabulary}{1.0\columnwidth}{C|C|C|C|C}
    \multirow{3}{*}{Model} & \multirow{3}{*}{Dataset} & \multicolumn{3}{c}{Accuracy (mIoU)} \\ \cline{3-5}
    & & \multirow{2}{*}{\shortstack[c]{Baseline\\(FPS)}} & \multirow{2}{*}{\shortstack[c]{Grid\\Sampling}} & \multirow{2}{*}{Diff.} \\ 
    & & & & \\
    \hline
    \hline
    \multirow{6}{*}{PV-L} 
    & \multirow{2}{*}{S3DIS} & \multirow{2}{*}{71.33} & \multirow{2}{*}{70.19} & \multirow{2}{*}{-1.14} \\
    & & & & \\
    \cline{2-5}
    & \multirow{2}{*}{ScanNet} & \multirow{2}{*}{70.70} & \multirow{2}{*}{69.84} & \multirow{2}{*}{-0.86} \\
    & & & & \\
    \cline{2-5}
    & \multirow{2}{*}{\shortstack[c]{Semantic\\KITTI}} & \multirow{2}{*}{50.91} & \multirow{2}{*}{51.66} & \multirow{2}{*}{+0.75} \\
    & & & & \\
    \hline
    \multirow{6}{*}{PMB-L} 
    & \multirow{2}{*}{S3DIS} & \multirow{2}{*}{69.72} & \multirow{2}{*}{69.16} & \multirow{2}{*}{-0.56} \\ 
    & & & & \\
    \cline{2-5}
    & \multirow{2}{*}{ScanNet} & \multirow{2}{*}{70.86} & \multirow{2}{*}{70.04} & \multirow{2}{*}{-0.82} \\
    & & & & \\
    \cline{2-5}
    & \multirow{2}{*}{\shortstack[c]{Semantic\\KITTI}} & \multirow{2}{*}{52.19} & \multirow{2}{*}{51.28} & \multirow{2}{*}{-0.91} \\ 
    & & & & \\
    \hline
    \end{tabulary}
}
\caption{Accuracy comparison of models using FPS and Grid Sampling in both training and inference.}
\label{table:with_training_accuracy}
\end{table}

\subsection{Latency Breakdown of MDPS}
\label{appendix:breakdown}
\begin{table}[t]
\setlength{\tabcolsep}{5pt}
\centering
{
    \small
    \begin{tabulary}{1.0\columnwidth}{C|C|C|C}
    \multirow{2}{*}{Task} & \multirow{2}{*}{S3DIS} & \multirow{2}{*}{ScanNet} & \multirow{2}{*}{\shortstack[c]{Semantic\\KITTI}} \\ 
    & & & \\
    \hline
    \hline
    \multirow{2}{*}{\shortstack[c]{Minimum Distance\\Curve Estimation}} & \multirow{2}{*}{48.52\%} & \multirow{2}{*}{52.39\%} & \multirow{2}{*}{39.48\%} \\
    & & & \\
    \hline
    \multirow{2}{*}{\shortstack[c]{Distance Curve\\Segmentation}} & \multirow{2}{*}{23.23\%} & \multirow{2}{*}{24.96\%} & \multirow{2}{*}{26.78\%} \\ 
    & & & \\
    \hline
    \multirow{2}{*}{\shortstack[c]{Sampling with\\Predicted Distance}} & \multirow{2}{*}{20.09\%} & \multirow{2}{*}{20.34\%} & \multirow{2}{*}{27.17\%} \\ 
    & & & \\
    \hline
    \multirow{2}{*}{Early Termination} & \multirow{2}{*}{8.16\%} & \multirow{2}{*}{2.31\%} & \multirow{2}{*}{6.57\%} \\ 
    & & & \\
    \hline
    \end{tabulary}
}
\caption{Latency breakdown of MDPS on various datasets.}
\label{table:latency_breakdown}
\end{table}

Table \ref{table:latency_breakdown} provides a detailed latency breakdown of four major operations involved in the MDPS: Minimum Distance Curve Estimation, Distance Curve Segmentation, Sampling with Predicted Distance, and Early Termination. For all datasets, Minimum Distance Curve Estimation accounts for the largest portion of the overall latency, ranging from approximately 40\% to 50\%. This is due to the high latency of initial FPS iterations required for estimation. Distance Curve Segmentation and Sampling with Predicted Distance each contribute a smaller portion of the latency (each 20-30\%), highlighting the effectiveness of increased parallelism in distance computation and reduced overhead in sampling. Lastly, Early Termination has the smallest portion, comprising less than 10\% of the overall latency. This indicates that significant overestimation rarely occurs, underscoring the high accuracy of our minimum distance curve estimator.

\subsection{Speedup of MDPS}
\label{appendix:sampling_speedup}

\begin{figure}[t]
    \centering
    \includegraphics[width=\columnwidth]{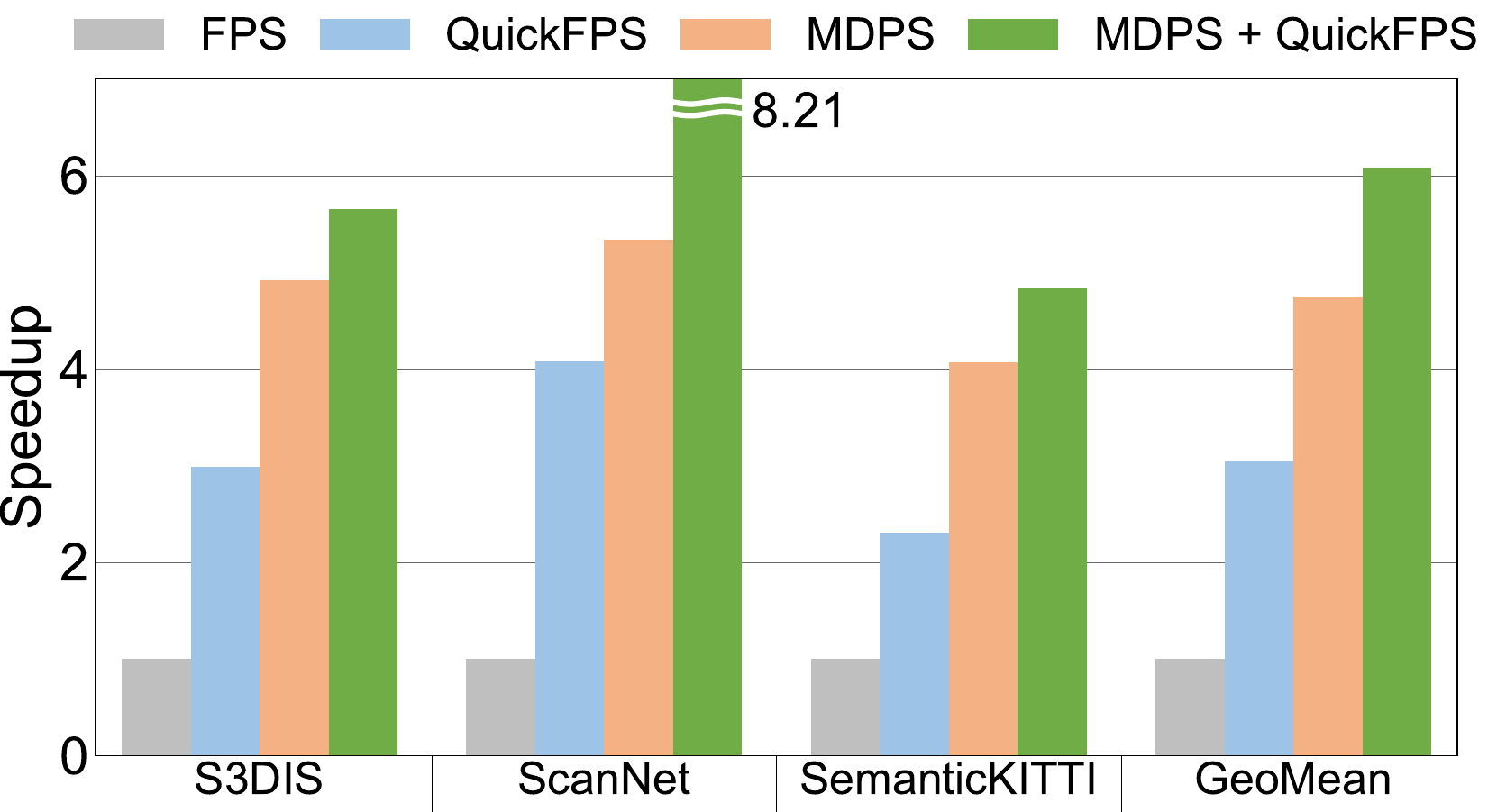}
    \caption{Sampling speedup of MDPS and QuickFPS.}
    \label{fig:sample_speedup}
\end{figure}

Figure~\ref{fig:sample_speedup} demonstrates the speedup specific to the sampling operation.
Across the S3DIS~\cite{s3dis}, ScanNet~\cite{scannet}, and SemanticKITTI~\cite{semantickitti} datasets, \ours\space achieves significant improvements in sampling latency, achieving a geomean speedup of 4.75$\times$ compared to FPS. When QuickFPS~\cite{quickfps, quickfps-sw} is integrated with \oursampling\space (i.e., FPS used in minimum distance curve estimation is replaced with QuickFPS), \oursampling\space achieves geomean speedup of 6.08$\times$ compared to baseline FPS and 2.00$\times$ compared to standalone QuickFPS. The high portion of minimum distance curve estimation in total \oursampling\space latency (Table~\ref{table:latency_breakdown}) explains the substantial improvement in sampling speed enabled by the integration of QuickFPS.

\subsection{Speedup of Redundancy Free Neighbor Search}
\label{appendix:ns_speedup}

\begin{figure}[t]
    \centering
    \includegraphics[width=0.9\columnwidth]{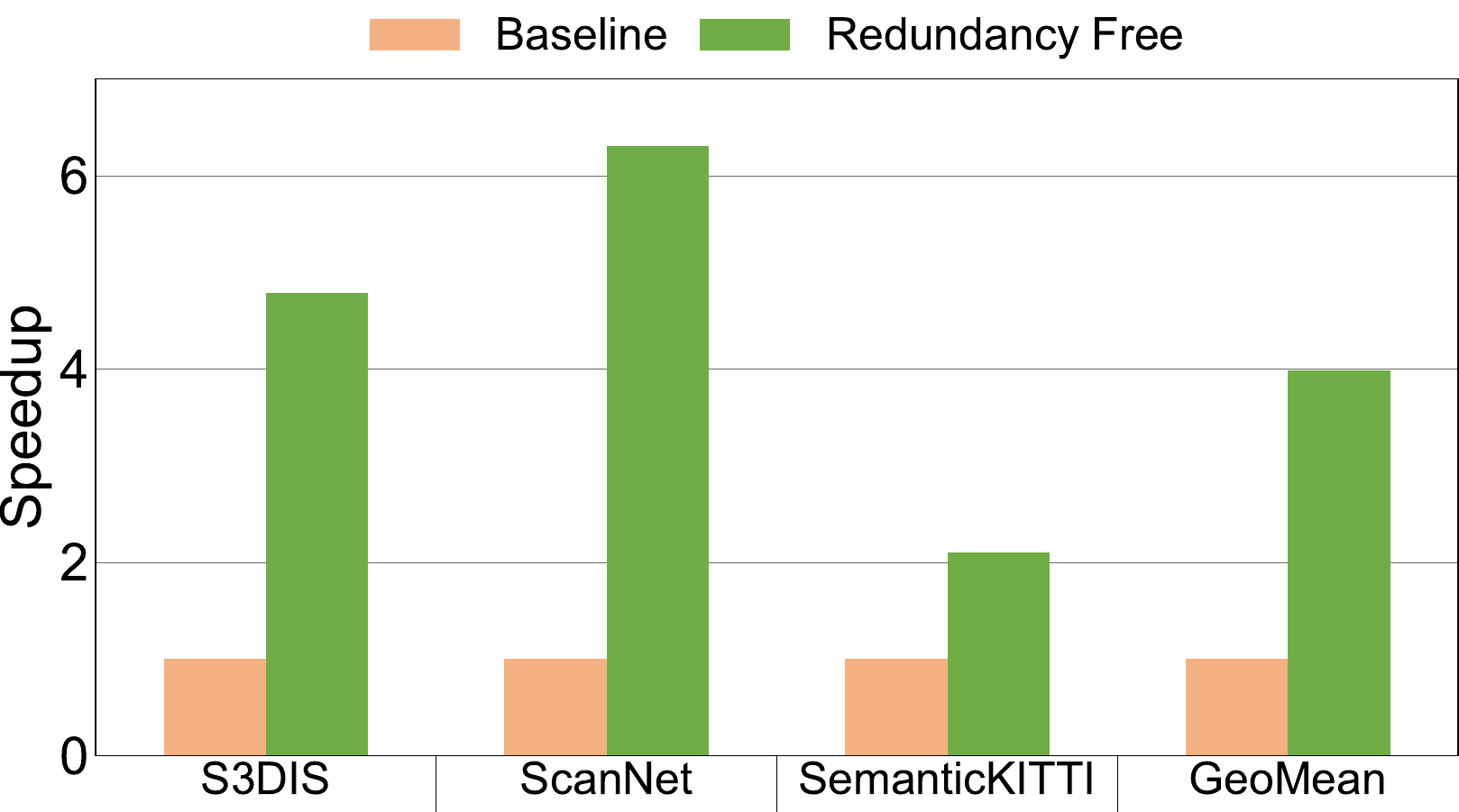}
    \caption{Speedup of Redundancy Free Neighbor Search.}
    \label{fig:ns_speedup}
\end{figure}

\begin{figure}[t]
    \centering
    \includegraphics[width=\columnwidth]{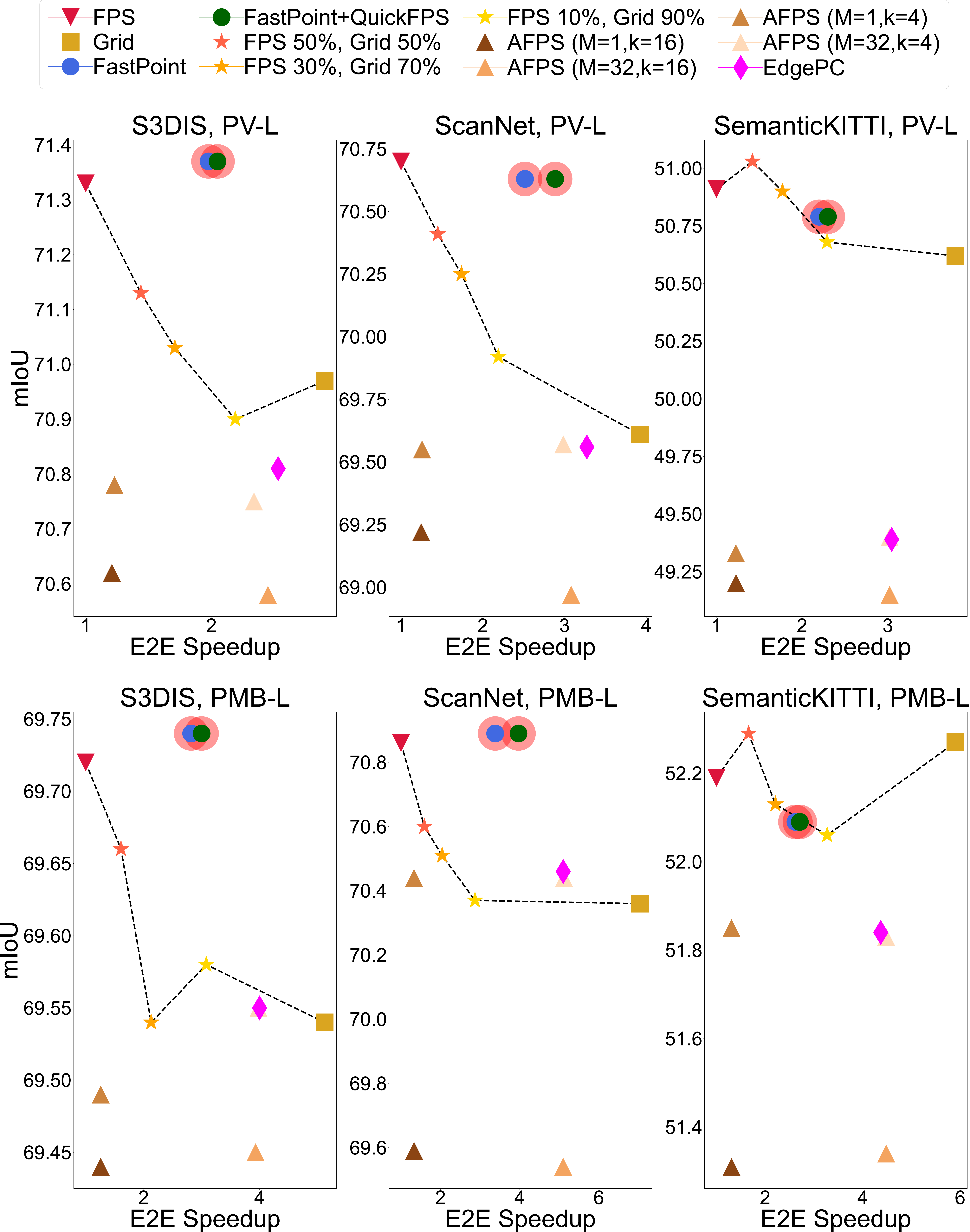}
    \caption{Speedup-mIoU curve of various sampling methods. PV and PMB stand for PointVector and PointMetaBase.}
    \label{fig:pareto}
\end{figure}


In this section, we report the speedup specific to the neighbor search operations (i.e., Ball Query, k-NN). Figure~\ref{fig:ns_speedup} highlights the significant latency reduction for neighbor search operations achieved through Redundancy Free Neighbor Search. Thanks to the reuse of exclusion list in Ball Query and search space reduction in k-NN, we achieve an impressive geomean speedup of 3.99$\times$ across the S3DIS, ScanNet, and SemanticKITTI datasets. These results confirm that FastPoint is not only effective in achieving speedup for sampling but also excels in accelerating neighbor search operations.


\subsection{Latency-Accuracy Comparison with Other Sampling Methods}
\label{appendix:pareto}
In this section, we compare latency and accuracy of \ours\space with other sampling methods. The speedup-mIoU plot (Figure~\ref{fig:pareto}) demonstrates that FastPoint lies above the Pareto front, while both Adjustable FPS~\cite{adjustable_fps} and EdgePC~\cite{edgepc} exhibit suboptimal performance. Although Grid sampling achieves higher speedup than FastPoint, it suffers notable loss in mIoU compared to FastPoint. For a more comprehensive comparison, we also evaluate a hybrid approach combining two existing state-of-the-art sampling algorithms: FPS and Grid sampling. To the best of our knowledge, FastPoint is the fastest sampling method that maintains both accuracy and sampling quality on par with FPS.

\subsection{Ablation Study for Initial FPS Iterations}
\label{appendix:ablation_init}
We perform an ablation study on the impact of the number of initial FPS iterations used for estimation. Table~\ref{tbl:ablation} demonstrates estimator error, sampling quality, model accuracy, and end-to-end speedup for different values of $p$, where $p$ represents the ratio of the initial FPS iterations to the total iterations. As $p$ decreases, end-to-end speedup improves due to reduced estimation time, while sampling quality and model accuracy decline due to the increased estimator error. To balance error and efficiency, we select $p=0.1$.

\begin{table}[t]
\centering
{\small
    \begin{tabulary}{1.0\columnwidth}{C|C|C|C}
     & $p$=0.2 & $p$=0.1 & $p$=0.05 \\
    \hline
    \hline
    Est. Error (\%) & 0.72 & 0.77 & 1.46 \\ \hline
    Smpl. Q. (\%) & 99.3 & 99.35 & 99.01 \\ \hline
    mIoU (\%) & 70.95 & 70.89 & 70.83 \\ \hline
    E2E Speedup & 2.70$\times$ & 3.38$\times$ & 3.82$\times$ \\ \hline
    \end{tabulary}
}
\caption{Ablation study for initial FPS iterations. PointMetaBase-L model and ScanNet dataset is used.}
\label{tbl:ablation}
\end{table}

\subsection{Robustness of FastPoint}
\label{appendix:robustness}
To evaluate the robustness of FastPoint against data augmentations, we apply the same training augmentations (i.e., jitter, rotation, scaling, and point dropping) to the validation set. As shown in Table~\ref{tbl:robustness}, these augmentations have minimal impact on estimator error, sampling quality, and model accuracy. This is because the minimum distance curve's smoothness is preserved despite the augmentations, which is the key of predictability. Considering that data augmentations used during training typically reflect the real-world variations, the experimental results substantiate the claim that FastPoint is robust in practical scenarios.

\begin{table}[t]
\centering
{\footnotesize
    \setlength{\tabcolsep}{5pt}
    \begin{tabulary}{1.0\columnwidth}{C|C|C|C|C|C|C}
     & \multicolumn{2}{c|}{S3DIS} & \multicolumn{2}{c|}{ScanNet} & \multicolumn{2}{c}{SemanticKITTI} \\\cline{2-7}
     & no aug & aug & no aug & aug & no aug & aug \\
    \hline
    \hline
    Est. Error (\%) & 1.02 & 0.98 & 0.77 & 0.93 & 1.93 & 2.05 \\ \hline
    Smpl. Q. (\%) & 99.45 & 99.40 & 99.35 & 99.10 & 98.36 & 98.30 \\ \hline
    mIoU diff (\%) & +0.02 & -0.03 & +0.03 & 0 & -0.1 & -0.11 \\ \hline
    \end{tabulary}
}
\caption{Robustness on augmentation. PointMetaBase-L is used when measuring mIoU.}
\label{tbl:robustness}
\end{table}

\begin{table}[t]
\centering
{\small
    \begin{tabulary}{1.0\columnwidth}{C|C|C|C}
     & \multirow{2}{*}{\shortstack[c]{ScanNet\\Estimator}} & \multirow{2}{*}{\shortstack[c]{S3DIS\\Estimator}} & \multirow{2}{*}{\shortstack[c]{SemanticKITTI\\Estimator}} \\ & & & \\
    \hline
    \hline
    Est. Error (\%) & 0.77 & 1.42 & 2.24 \\ \hline
    Smpl. Q. (\%) & 99.35 & 99.28 & 98.89 \\ \hline
    mIoU (\%) & 70.86 & 70.92 & 70.93 \\ \hline
    \end{tabulary}
}
\caption{Cross-dataset applicability of \ours. Experiments are performed on PointMetaBase-L model, ScanNet dataset.}
\label{tbl:applicability}
\end{table}

\subsection{Cross-Dataset Applicability of \ours}
\label{appendix:applicability}
To explore the cross-dataset applicability of estimator, we apply estimator trained on S3DIS and SemanticKITTI dataset to ScanNet dataset. The estimator trained on S3DIS performed well on ScanNet, while the SemanticKITTI estimator had a relatively higher error and lower sampling quality, indicating better compatibility within similar indoor datasets but challenges in cross-environment generalization. Still, the sampling quality loss is minimal ($<2\%$) in all cases, demonstrating no impact on model accuracy.

\subsection{Visualizing Minimum Spacing Distribution}
\label{appendix:visualization}
\begin{figure}[t]
    \centering
    \begin{subfigure}[t]{1\linewidth}
        \centering
        \includegraphics[width=0.8\linewidth]{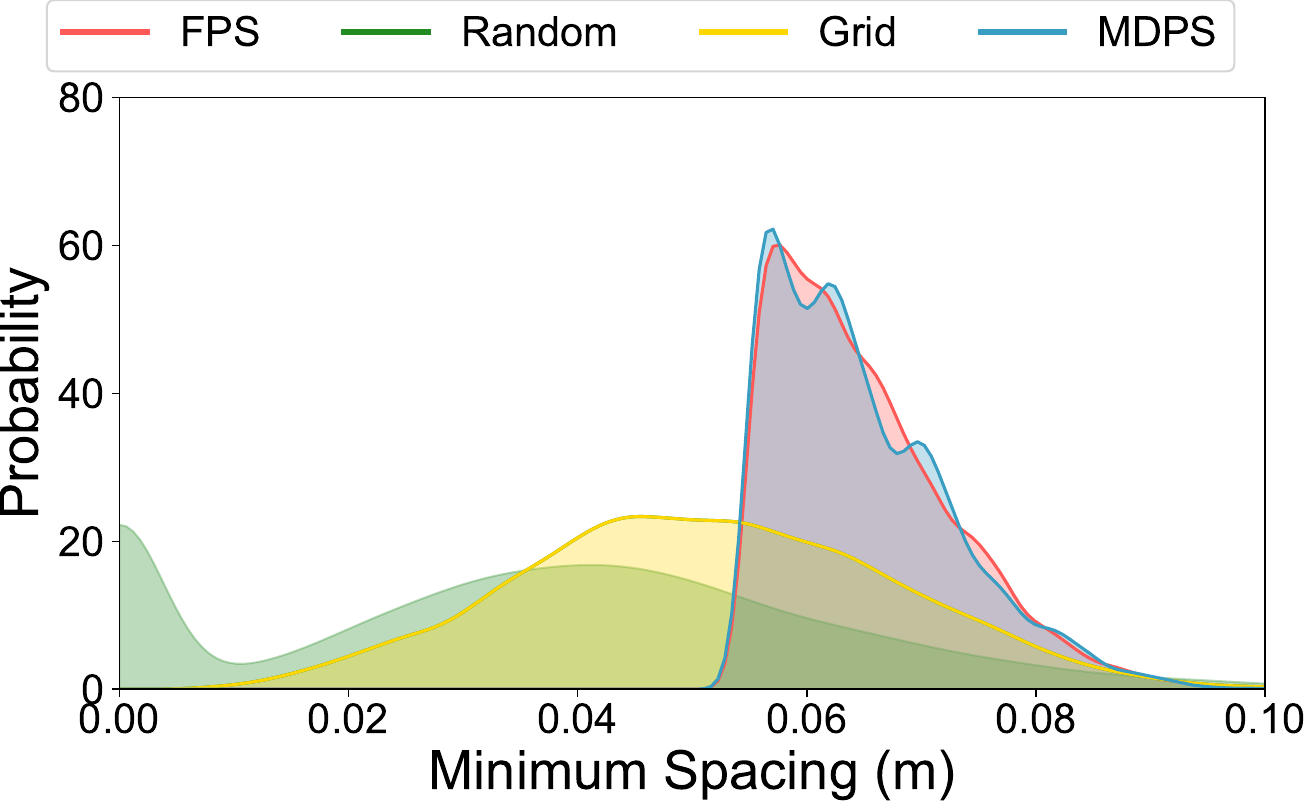}
        \caption{Minimum spacing distribution for S3DIS.}
        \label{fig:distribution_S3DIS}
    \end{subfigure}
    
    \begin{subfigure}[t]{1\linewidth}
        \centering
        \includegraphics[width=0.8\linewidth]{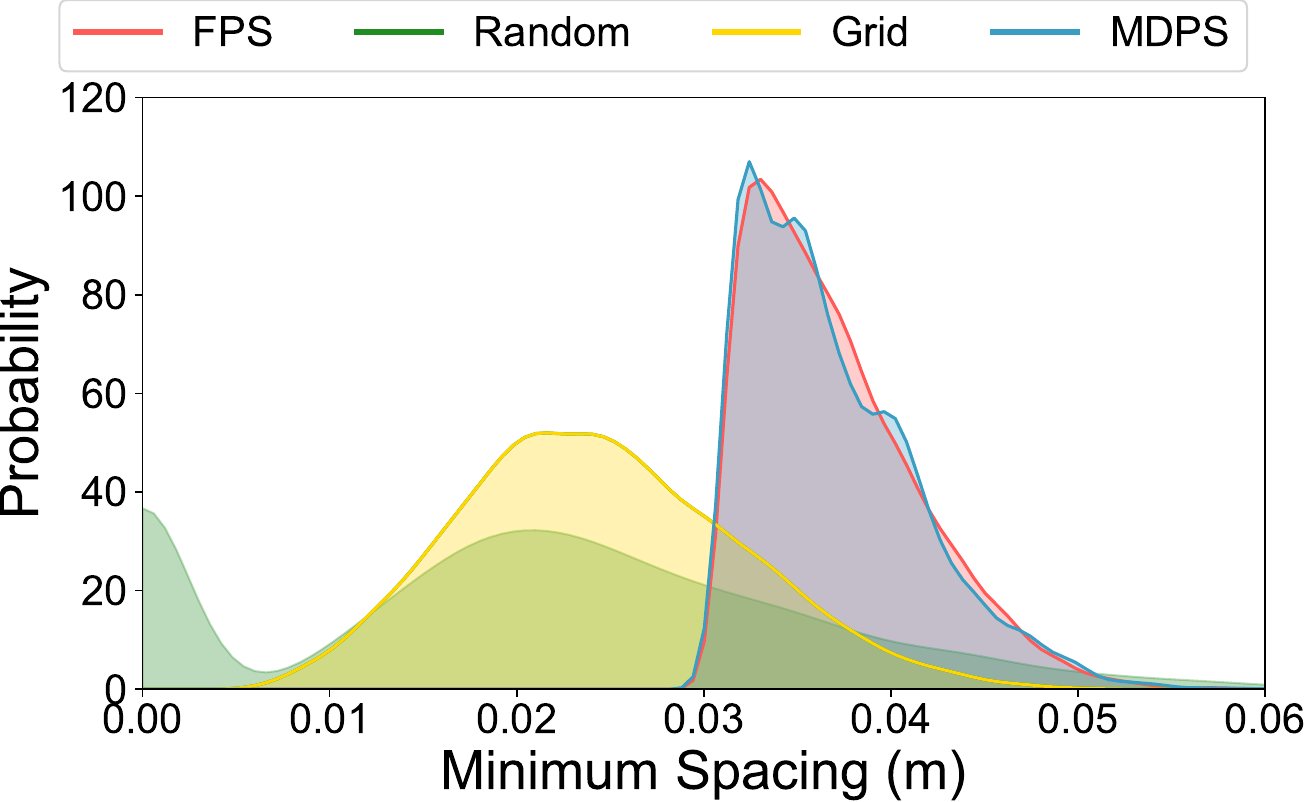}
        \caption{Minimum spacing distribution for ScanNet.}
        \label{fig:distribution_ScanNet}
    \end{subfigure}
    
    \begin{subfigure}[t]{1\linewidth}
        \centering
        \includegraphics[width=0.8\linewidth]{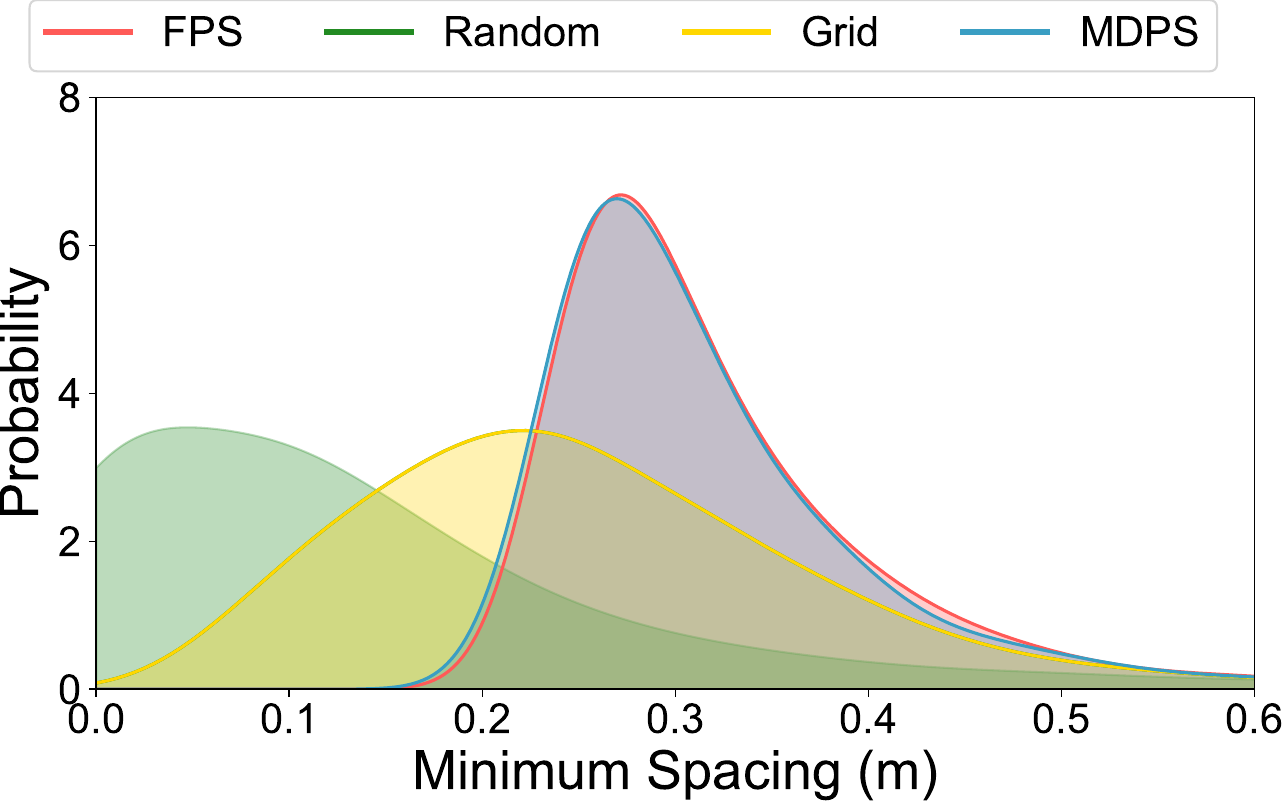}
        \caption{Minimum spacing distribution for SemanticKITTI.}
        \label{fig:distribution_SemanticKITTI}
    \end{subfigure}

    \caption{Minimum spacing distribution of each sampling method for S3DIS, ScanNet, and SemanticKITTI.}
    \label{fig:distribution}
\end{figure}

Figure~\ref{fig:distribution} illustrates the distribution of minimum spacing between sampled points using FPS, Random Sampling, Grid Sampling, and \oursampling. \oursampling\space demonstrates a distribution highly similar to that of FPS, confirming its effectiveness in resembling the high sampling quality of FPS. In contrast, Grid Sampling and Random Sampling exhibit distributions that deviate significantly from FPS, indicating lower sampling quality.

\subsection{Scalability to Non-PointNet++-Based Models}
\label{appendix:nonpointnet}
FastPoint is a \emph{model-agnostic technique} accelerating FPS and neighbor search. It is applicable to any point cloud models that use these operations. We applied FastPoint to Point Transformer model adopting FPS~\cite{ptr}, achieving 2.16$\times$ end-to-end speedup without sacrificing accuracy as shown in Table~\ref{tbl:applicability}.
Furthermore, several recent Mamba-based models with FPS~\cite{pointcloudmamba, pointmamba} achieve state-of-the-art performance on various datasets, substantiating that FastPoint has the potential to accelerate a wide range of emerging point cloud models.

\begin{table}[t]
\centering
{\footnotesize
    \setlength{\tabcolsep}{5pt}
    \begin{tabulary}{1.0\columnwidth}{C|C|C|C|C|C|C}
    & FPS & Grid & FastPoint & FastPoint+QuickFPS \\\cline{2-5}
    \hline
    \hline
    mIoU (\%) & 70.85 & 70.34 & 70.74 & 70.74 \\ \hline
    E2E speedup & 1$\times$ & 2.97$\times$ & 2.16$\times$ & 2.27$\times$ \\ \hline
    \end{tabulary}
}
\caption{Speedup and mIoU on Point Transformer, S3DIS dataset.}
\label{tbl:applicability}
\end{table}

\end{document}